\documentclass[sigconf]{acmart}
\usepackage{algorithm}
\usepackage{algorithmic}
\usepackage{multirow}
\usepackage{booktabs}
\usepackage{colortbl}
\usepackage{diagbox}
\usepackage{subfigure}
\usepackage{balance}

\AtBeginDocument{%
  \providecommand\BibTeX{{%
    \normalfont B\kern-0.5em{\scshape i\kern-0.25em b}\kern-0.8em\TeX}}}

\copyrightyear{2022}
\acmYear{2022}
\setcopyright{acmcopyright}\acmConference[MM '22]{Proceedings of the 30th ACM International Conference on Multimedia}{October 10--14, 2022}{Lisboa, Portugal}
\acmBooktitle{Proceedings of the 30th ACM International Conference on Multimedia(MM '22), October 10--14, 2022, Lisboa, Portugal}
\acmPrice{15.00}
\acmDOI{10.1145/3503161.3548062}
\acmISBN{978-1-4503-9203-7/22/10}

\begin{document}



\title{Exploring Effective Knowledge Transfer for Few-shot Object Detection}



\author{Zhiyuan Zhao$^{1}$}
\affiliation{
  \institution{State Key Laboratory of Virtual Reality Technology and Systems, Beihang University}
  \city{Beijing}
  \country{China}
}
\email{zhaozhiyuan@buaa.edu.cn}

\author{Qingjie Liu$^{2}$}
\authornote{Corresponding Author.}
\affiliation{
  \institution{State Key Laboratory of Virtual Reality Technology and Systems, Beihang University}
  \city{Beijing}
  \country{China}
}
\email{qingjie.liu@buaa.edu.cn}

\author{Yunhong Wang$^{3}$}
\affiliation{
  \institution{State Key Laboratory of Virtual Reality Technology and Systems, Beihang University}
  \city{Beijing}
  \country{China}
}
\email{yhwang@buaa.edu.cn}




\renewcommand{\shortauthors}{Zhiyuan Zhao, Qingjie Liu, \& Yunhong Wang}

\begin{abstract}

Recently, few-shot object detection~(FSOD) has received much attention from the community, and many methods are proposed to address this problem from a knowledge transfer perspective. Though promising results have been achieved, these methods fail to achieve shot-stable:~methods that excel in low-shot regimes are likely to struggle in high-shot regimes, and vice versa. We believe this is because the primary challenge of FSOD changes when the number of shots varies. In the low-shot regime, the primary challenge is the lack of inner-class variation. In the high-shot regime, as the variance approaches the real one, the main hindrance to the performance comes from misalignment between learned and true distributions. However, these two distinct issues remain unsolved in most existing FSOD methods. In this paper, we propose to overcome these challenges by exploiting rich knowledge the model has learned and effectively transferring them to the novel classes. For the low-shot regime, we propose a distribution calibration method to deal with the lack of inner-class variation problem. Meanwhile, a shift compensation method is proposed to compensate for possible distribution shift during fine-tuning. For the high-shot regime, we propose to use the knowledge learned from ImageNet as guidance for the feature learning in the fine-tuning stage, which will implicitly align the distributions of the novel classes. Although targeted toward different regimes, these two strategies can work together to further improve the FSOD performance. Experiments on both the VOC and COCO benchmarks show that our proposed method can significantly outperform the baseline method and produce competitive results in both low-shot settings~(shot<5) and high-shot settings~(shot$\ge$5). Code is available at \href{https://github.com/JulioZhao97/EffTrans\_Fsdet.git.}{https://github.com/JulioZhao97/EffTrans\_Fsdet.git.}

\end{abstract}






\maketitle

\begin{figure*}[t]
    \centering
    \includegraphics[width=1.0\textwidth]{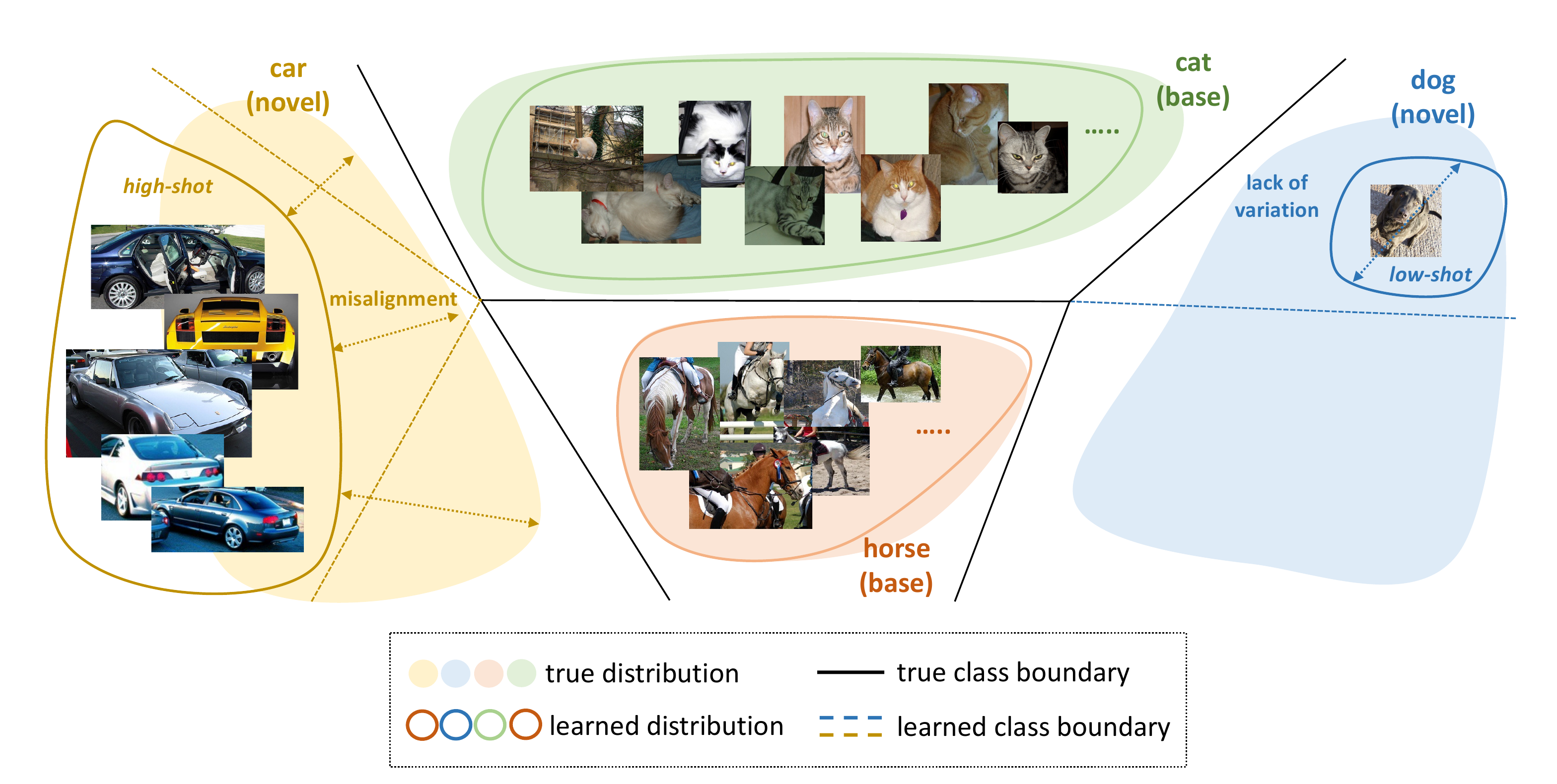}
    \caption{An illustration of the cause of shot-unstable. This figure depicts a high-shot case~(shot=5) and a low-shot case~(shot=1). We believe as the number of shot increases, the main challenges of FSOD also changes. (1)~In the low-shot regime, the key issue of FSOD is the lack of variation in novel classes. (2)~In the high-shot regime, the main challenge changes to the distribution misalignment.}
    \label{figure1}
\end{figure*}

\section{Introduction}

Today, deep neural networks~(DNNs) have achieved outstanding performance in a wide variety of data-intensive applications. Unfortunately, due to the data-driven nature of deep models, their performance is severely hampered in data-limited scenarios. On the other hand, humans have an incredible capacity for learning from a few examples and generalizing to new concepts from a small amount of information. For instance, by showing a photograph of a stranger to a child once, he/she can rapidly identify this stranger from a pile of pictures. To acquire this ability, researchers endeavor to empower the deep models with the quick and robust learning ability in data-limited scenarios, termed few-shot learning~\cite{DBLP:conf/mm/WangWLL21,DBLP:conf/mm/LiuMZYL21,DBLP:conf/mm/DongYXHY21,DBLP:conf/mm/0001YC21}. 

As a fundamental task in computer vision, object detection has witnessed tremendous progress in the last few years, yet it still suffers from the data curse. Precedented by few-shot recognition, efforts have been devoted to addressing few-shot object detection~(FSOD), which is a much more challenging task. Earlier attempts~\cite{DBLP:conf/aaai/ChenWW018} inherit ideas from approaches for few-shot classification~\cite{DBLP:conf/cvpr/RazavianASC14,DBLP:conf/icml/DonahueJVHZTD14} and adapt them to FSOD. For instance, following the meta-learning paradigm~\cite{DBLP:conf/nips/VinyalsBLKW16,DBLP:conf/nips/SnellSZ17}, meta-detectors~\cite{DBLP:conf/iccv/YanCXWLL19,DBLP:conf/iccv/YanCXWLL19,DBLP:conf/iccv/KangLWYFD19} are trained on the base classes to learn prior knowledge of the base classes and then are updated on the novel classes to make predictions. Another line of work involves the fine-tuning framework. Fine-tuning based FSOD methods normally consist of two steps:~(1)~firstly, base detectors are pre-trained on abundant base classes; (2) secondly, the detectors are fine-tuned on novel classes for adaption. These methods intend to transfer knowledge learned on base classes to the novel classes and are known as transfer learning based methods. However, due to the rarity of the target data, fine-tuning all parameters of the models is inefficient. Wang et al.~\cite{DBLP:conf/icml/WangH0DY20} propose to fine-tune only the classification and regression branches of the detector while freezing the feature extractor. Their approach yields competitive results with this simple strategy and reinvigorates researchers' interest in transfer learning methodology~\cite{DBLP:conf/cvpr/ZhangW21a,DBLP:conf/cvpr/FanMLS21}.

Despite the progress made, the performance of existing transfer learning based FSOD methods are still far from satisfying. We notice that most of these methods that perform well in low-shot regimes are likely to be inferior in high-shot regimes, and vice versa. In other words, these methods are incapable of achieving shot-stable. We believe that this is because the primary challenges in the low-shot regime and the high-shot regime are very different from each other~(as shown in Figure~\ref{figure1}). In the low-shot regimes, the primary difficulty is the lack of inner-class variation. In the high-shot regimes, to improve the detection performance, the key factor is to tackle misalignment between learned distributions and true distributions. However, in most existing transfer learning based methods, these two distinct issues are ignored thus their performance is unstable across different shots. 


To overcome these unsolved issues, a key point is to exploit the rich knowledge the model has learned in previous stages. To this end, we propose a calibration and regularization based method that enables effective knowledge transfer. Specifically, we propose a distribution calibration method to deal with the lack of variation issue for the low-shot regime. We calibrate the biased novel class distributions with the base class distributions. Then synthetic training features are sampled from calibrated novel distributions and added to training subsequently. In this calibration-and-generation manner, the inner-class variation of novel classes can be greatly enriched. Besides, we notice that there exist distribution shift of base classes due to the fine-tuning process, which may mislead novel class distributions and limit the effectiveness of calibration. To overcome this limitation, we present a strategy of compensating for possible distribution shift, namely shift compensation. 

In the high-shot regime, we propose to use the knowledge learned from ImageNet as guidance for feature learning in the fine-tuning stage, which will implicitly align the distributions of the novel classes. This is inspired by the recent studies that ImageNet features are stable and expressive and thus can be used as teachers for down-stream task learning~\cite{DBLP:journals/corr/abs-2111-14887,DBLP:conf/iccv/QiaoZLQWZ21}. We have tried the base detector model as the teacher, which also adopts the ImageNet pre-trained backbone, however, receiving unsatisfactory results. We suspect the ImageNet features are corrupted by the base class training. In this paper, the knowledge transfer from ImageNet features is achieved with a regularization loss. Although the two solutions are targeted for different regimes of FSOD, they can be combined together for performance boost. Our contributions are as follows:


(1)~We investigate the fundamental cause of the shot-unstable problem of FSOD: the key issue for FSOD in the low-shot regime and high-regime are different from each other and needs targeted solution. 

(2)~We propose two effective knowledge transfer strategies targeted for the low-shot and high-shot regimes of FSOD, respectively. The two strategies can be combined together for further improvement of FSOD.


(3)~Experiments on VOC and COCO benchmarks show that our method significantly outperforms the baseline method and achieves competitive results in both low-shot (shot<5) and high-shot (shot$\ge$5) regimes.

\section{Related Work}

\subsection{Object Detection}
Modern object detectors are built on top of deep neural networks. These methods can be broadly divided into single-stage detectors and two-stage detectors. Single-stage detectors are usually with high detection efficiency however a relatively low detection accuracy~\cite{DBLP:conf/eccv/LiuAESRFB16,DBLP:journals/corr/RedmonDGF15,DBLP:conf/iccv/LinGGHD17,DBLP:journals/corr/abs-1804-02767}. Two-stage detectors mostly refer to Faster-RCNN~\cite{DBLP:conf/nips/RenHGS15} and its derivatives~\cite{DBLP:journals/corr/abs-1805-00500,DBLP:journals/corr/abs-1712-00726,DBLP:conf/cvpr/0008CYLWL020,DBLP:conf/cvpr/PangCSFOL19}. They usually attain higher performance than single-stage detectors thanks to the stage-wise refining pipeline. Also, this flexible architecture makes them easily adaptable to extended tasks such as FSOD. In addition to these two families, some anchor-free detection methods~\cite{DBLP:conf/iccv/TianSCH19,DBLP:conf/iccv/YangLHWL19,DBLP:conf/eccv/LawD18,DBLP:conf/cvpr/ZhouZK19,DBLP:conf/cvpr/ZhuHS19} are proposed to release detectors from burdensome anchor settings.

\subsection{Few-shot Object Detection}
Few-shot object detection approaches can be broadly grouped into three branches: transfer learning based methods, metric-learning based methods, and meta-learning based methods. Transfer learning based methods mainly focus on transferring knowledge from base classes to novel classes and unleashing the potential of fine-tuning~\cite{DBLP:conf/iccv/YanCXWLL19,DBLP:conf/iclr/0007WKWA21,DBLP:conf/iclr/LeeCK20}. Chen et al.~\cite{DBLP:conf/aaai/ChenWW018} combine the advantages of Faster RCNN and SSD to alleviate the transfer difficulties from the source domain to the target domain. Recently, Wang et al.~\cite{DBLP:conf/icml/WangH0DY20} rekindle the interest in transfer learning by showing its potential in improving FSOD, which inspires a lot of follow-up works~\cite{DBLP:conf/cvpr/SunLCYZ21,DBLP:conf/cvpr/ZhangW21a}. Their proposed method, named TFA, freezes the parameters of the model trained on the base classes and only fine-tunes the detection head with the novel classes. Meta-learning based methods follow the ideas of meta-learning in few-shot classification task~\cite{DBLP:conf/nips/VinyalsBLKW16,DBLP:conf/nips/SnellSZ17,DBLP:conf/nips/OreshkinLL18,DBLP:conf/icml/FinnAL17,DBLP:conf/iclr/RaviL17,DBLP:conf/nips/SchwartzKSHMKFG18,DBLP:conf/iclr/RenTRSSTLZ18,DBLP:conf/cvpr/LeeMRS19}, which intend to learn generic knowledge across base classes and then generalize to novel classes. Specifically, Yan et al.~\cite{DBLP:conf/iccv/YanCXWLL19} propose to conduct meta-learning over RoI regions. Kang et al.~\cite{DBLP:conf/iccv/KangLWYFD19} design a few-shot detection method using a meta feature learner and a reweighting module. 

Metric-learning based methods focus on learning better representations~\cite{DBLP:journals/jmlr/SalakhutdinovTT12,koch2015siamese,DBLP:conf/cvpr/SungYZXTH18,DBLP:conf/eccv/ZhaoZYF18}. Leonid et al.~\cite{DBLP:conf/cvpr/KarlinskySHSAFG19} propose a sub-network to learn an embedding space and apply it for novel class detection. Sun et al.~\cite{DBLP:conf/cvpr/SunLCYZ21} adopt supervised contrastive learning to learn better representations, which is implemented with an additional contrastive branch to guide RoI feature learning. Except for these methods, there are some other interesting works proposed for FSOD.  Zhang et al.~\cite{DBLP:conf/cvpr/ZhangW21a} 's work demonstrates that hallucination is helpful for few-shot detection. Wu et al.~\cite{DBLP:journals/corr/abs-2103-01077} believe that there is a universal prototype across all categories, with which the features learned from base classes can be generalized well to novel classes. Qiao et al.~\cite{DBLP:conf/iccv/QiaoZLQWZ21} propose DeFRCN, which largely improves few-shot detection performance through multi-stage and multi-task decoupling.

\subsection{Distribution Calibration}

The goal of Distribution Calibration~(DC) is to align a target distribution to a reference distribution. The idea has been used for solving many unbalanced distribution problems. In ~\cite{DBLP:conf/cvpr/ZhangLY0S21}, Zhang et al. investigate the performance bottleneck of the
two-stage learning framework and proposed a unified distribution alignment strategy for long-tail visual recognition. Distribution calibration is also adopted in regression tasks~\cite{DBLP:conf/icml/SongDKF19}. Their work shows that predictions from previously trained regression models can be improved through distribution calibration. In \cite{DBLP:conf/cvpr/ShenLQ0CS21}, Shen et al. propose to learn a binary network by calibrating the latent representation through a teacher-student paradigm. Recently, Yang et al.~\cite{DBLP:conf/iclr/YangLX21} propose to reuse the statistics from many-shot classes and transfer them to better estimate the distributions of the few-shot classes according to their class similarities.


\begin{figure}[t]
    \centering
    \includegraphics[width=\columnwidth]{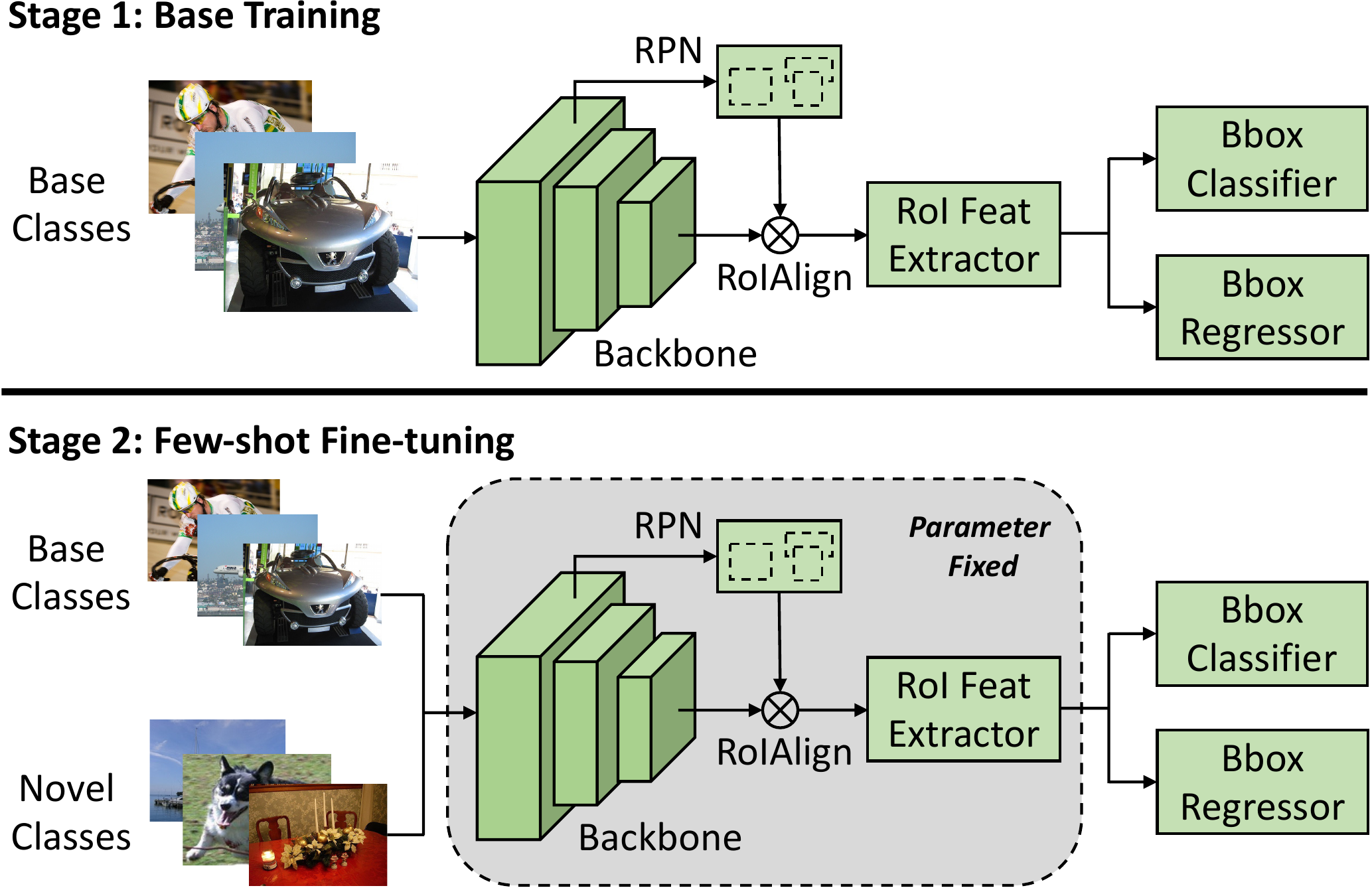}
    \caption{Illustration of baseline method TFA. The learning procedure of TFA consists of 2 stages:~(1)~base training and (2)~few-shot fine-tuning. During fine-tuning, only parameters of the detection head are updated.}
    \label{fig:fig2}
\end{figure}

\section{Our Method}
Before delving into the details of our proposed method, we first review the basic problem setting of FSOD and then take a brief look at our baseline method TFA.

\begin{figure*}[ht]
    \centering
    \includegraphics[width=1.0\textwidth]{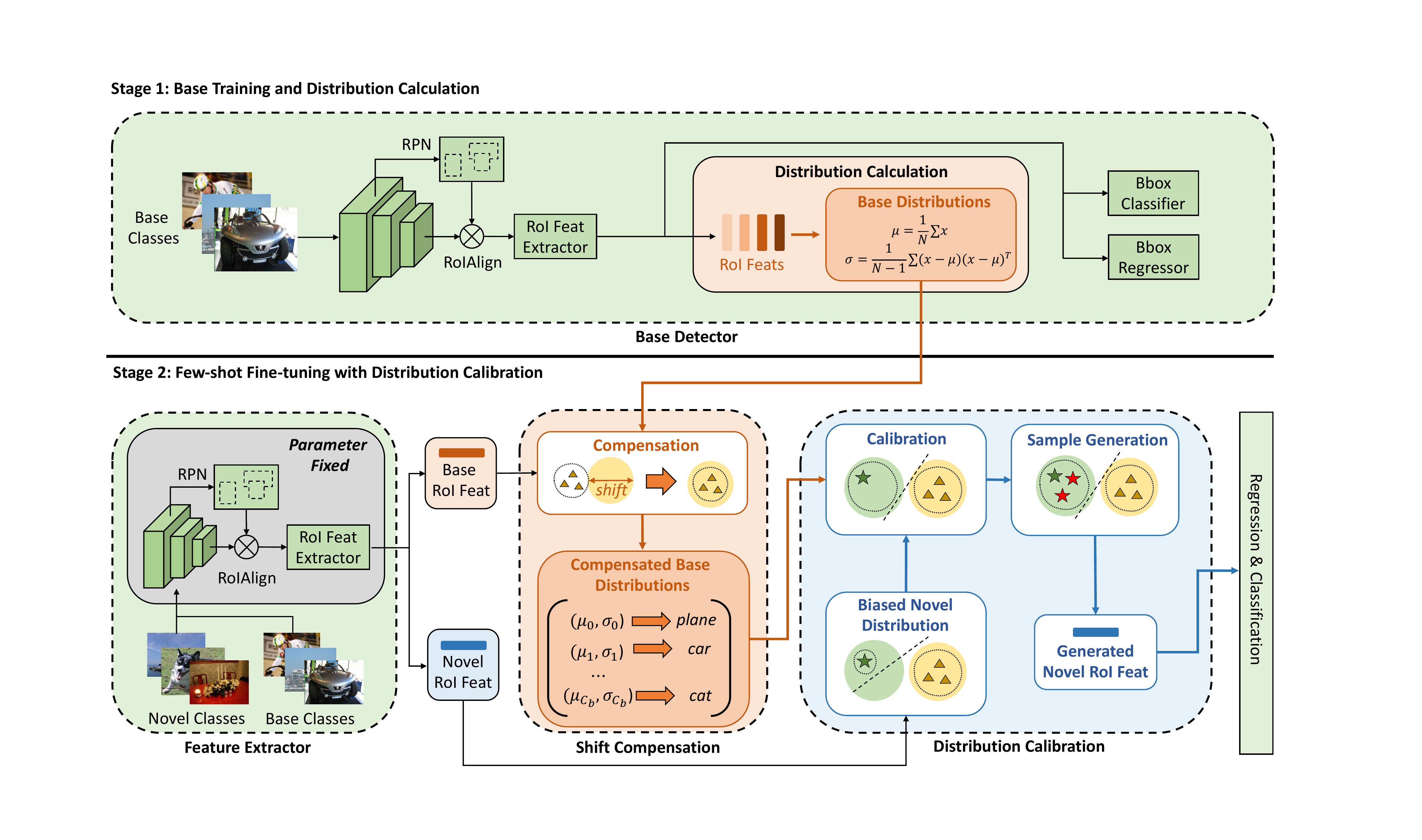}
    \caption{Illustration of our proposed distribution calibration. Our method is built on top of TFA. Similar to TFA, we train a Faster-RCNN on base classes and compute statistic information of base classes. In the second stage, we freeze the backbone and update the detection head with distribution calibration. Firstly, shift compensation is applied to compensate for potential base distributions shift. Then, biased novel distributions are calibrated by compensated base distributions. Finally, we draw samples from the calibrated novel distributions for training the detection head. In this manner, the lack of variation issue in low-shot settings is alleviated.}
    \label{fig1}
\end{figure*}

\subsection{Problem Setting}
We follow the FSOD setting introduced in \cite{DBLP:conf/iccv/KangLWYFD19}. Given two sets of classes: base classes $C_b$ and novel classes $C_n$, the learning procedure of few-shot detection is normally divided into two stages. The first stage is training a base model on sufficient training data of base classes $C_b$. On the second stage, model is fine-tuned on novel classes $C_n$. $K$ samples~(normally $K\le10$) for each class are used in the second stage fine-tuning. It is worth noting that there is no overlap between $C_b$ and $C_n$, that is $C_b \cap C_n = \varnothing$. In the second stage, to preserve the performance on base classes, model is fine-tuned on a balanced set containing training samples from both base classes $C_b$ and novel classes $C_n$.

\subsection{Review of TFA}
TFA is built on top of a two-stage detector Faster-RCNN. It consists of two learning stages. In the first stage, the Faster-RCNN is trained on base classes $C_{b}$. In the second stage, the Faster-RCNN model is fine-tuned on novel classes $C_n$, where each class contains $K$ training samples. Notably, during the fine-tuning stage, parameters of the backbone network and RoI feature extractor are fixed, only parameters of the prediction head, i.e., the classification branch and the regression branch are fine-tuned. The two-stage learning procedure and model construction of TFA are illustrated in Figure~\ref{fig:fig2}. The essential idea of TFA is that it provides a shared backbone for base and novel classes, which makes it feasible to transfer knowledge from base classes to novel classes. The only component in the detection model we should concern with is the task-relevant prediction head, and we fine-tune it to adapt the model to the novel classes. 

\subsection{Base Training and Distribution Calculation}

In this section, we present our proposed method in detail. The learning procedure of our proposed method is divided into two steps: (1) base training and distribution calculation and (2) few-shot fine-tuning on novel classes with distribution calibration and distribution regularization. In the first stage, a base detector (e.g., Faster-RCNN) is trained on base classes $C_b$. And then, we calculate base class distributions used for the subsequent distribution calibration.

\noindent
\textbf{Distribution Calculation.}
In a Faster-RCNN model, RoI feature $x^c_i \in R^d$ is extracted by the RoI feature extractor for each proposal. Then we use them to calculate the distributions of all base classes. For simplicity, we assume that the class distribution $\Omega_c$ of each category is a multidimensional Gaussian distribution. Formally, it can be written as $\{ \Omega_c=(\mu_c, \sigma_c)|c \in C_b, \mu_c \in R^d, \sigma_c \in R^{d \times d}\}$. $\mu_c$ and $\sigma_c$ are mean and covariance of $\Omega_c$. We use all RoI features that are considered as positive samples~(IOU$\ge$0.5) to calculate the distributions of the base classes. The mean and variance are calculated as follows~($c \in C_b$ denotes a base class, $N_c$ denotes the number of positive RoI features of class $c$ in calculation):

\begin{equation}
    \mu_c = \frac{1}{N_c} \sum_{i}^{N_{c}} x^c_i
\end{equation}
\begin{equation}
    \sigma_c = \frac{1}{N_c-1} \sum_{i}^{N_{c}} (x^c_i- \mu_c)(x^c_i- \mu_c)^ \mathrm{ T }
\end{equation}


\subsection{Distribution Calibration}

In this section, we present our proposed distribution calibration method targeted at the low-shot regime. In distribution calibration, biased novel classes distributions are calibrated using distributions of the base classes, which are learned by sufficient samples. Meanwhile, shift compensation is proposed to obtain a better estimation of base class distributions.


\noindent
\textbf{Shift Compensation.}
After the first stage training, we have a detector well-trained on abundant base class data and the distributions of the base classes. In the second stage, the detection model is fine-tuned on novel classes. Considering that when fine-tuning the model on new categories~(novel classes), the distributions of base classes may shift~\cite{DBLP:conf/cvpr/0004TLHWCJ020,DBLP:conf/cvpr/LiYLLJY21}. Therefore, we propose to compensate for this shift to make sure that the base class distributions would not mislead the novel class distributions.

Formally, we use $t$ to denote the current training iteration. All base class positive RoI features at iteration $t$ are denoted as $X^{t} = \{ x_i^{t} \}$ and base distributions are denoted as $\{\Omega_c=(\mu_c,\sigma_c)\}_{c=1}^{C_b}$. For each positive base class RoI features $x_i^t$ in $X^{t}$~(class label of $x_i^t$ is denoted as $c$), distribution of base class $c$ is compensated as follows:

\begin{equation}
    \mu_c = \theta \times \mu_c + (1-\theta) \times x_i^t
\end{equation}

\noindent
where $\theta$ is a hyper-parameter controlling the compensation degree. 

\noindent
\textbf{Distribution Calibration.}
After compensation, we use the compensated base class distributions to calibrate the biased novel class distributions. Specifically, for a novel RoI feature $x$ of class $n$ in the fine-tuning stage, we find its $k$ nearest base classes and use them to calibrate the novel distribution. Formally, we have:

\begin{equation}
    D = \{d_c = ||x - u_c||_2, c \in C_b \}
\end{equation}
\begin{equation}
    C_b^{cal} = \{c|d_c \in \min_k D, c \in C_b \}
\end{equation}

\noindent
where $C_b^{cal}$ stands for the $k$ nearest base classes. The closer the centers of two classes to each other, the more similar they are. Therefore, the knowledge is more likely to be shared between them. Here, the biased distribution of novel class $n$ is calibrated with its $k$ nearest base classes as follows:

\begin{equation}
    \mu_n = x + \frac{1}{k} \sum_{c}^{|C_b^{cal}|} \mu_c
\end{equation}

\begin{equation}
    \sigma_n = \frac{1}{k} \sum_{c}^{|C_b^{cal}|} \sigma_c + \alpha
\label{e6}
\end{equation}


\noindent
$\mu_n$ and $\sigma_n$ denotes the calibrated novel distribution of class $n$. $\alpha$ is a hyper-parameter which controls the dispersion of the calibrated distribution. After calibration, we sample $M$ synthetic RoI features $\{ \tilde{x} \}$ from the calibrated novel distribution. This procedure can be denoted as:


\begin{equation}
    \{ \tilde{x} \} = \{ \tilde{x} | \tilde{x} \sim \Omega_n=(\mu_n, \sigma_n) \}
\end{equation}

Finally , we feed both real sample $x$ and synthetic samples $\{ \tilde{x} \}$ into the detection head for training. The loss values in the fine-tuning stage come from both real and synthetic samples. Benefit from this calibration-and-generation process, the inner-class variation of the novel classes is enriched, and the over-fitting will be alleviated.

\begin{table*}[t]
    \centering
    \begin{tabular}{c|l|ccccc|ccccc|ccccc}
        \toprule 
        \multirow{2}{*}{Type} & 
        \multirow{2}{*}{Method} & 
         \multicolumn{5}{c|}{Split1} & \multicolumn{5}{c|}{Split2} & \multicolumn{5}{c}{Split3}\\
        & & 1 & 2 & 3 & 5 & 10
        & 1 & 2 & 3 & 5 & 10
        & 1 & 2 & 3 & 5 & 10\\
        \midrule
        \multirow{6}{*}{\shortstack{Meta \\ Learning}} & FSRW~\cite{DBLP:conf/iccv/KangLWYFD19}$_{ICCV19}$ &14.8 &15.5 &26.7 &33.9 &47.2
        &5.7 &15.3 &22.7 &30.1 &40.5
        &21.3 &25.6 &28.4 &42.8 &45.9\\
        & MetaDet~\cite{DBLP:conf/iccv/WangRH19}$_{ICCV19}$ 
        & 18.9 & 20.6 & 30.2 & 36.8 & 49.6
        & 21.8 & 23.1 & 27.8 & 31.7 & 43
        & 20.6 & 23.9 & 29.4 & 43.9 & 44.1\\
        & Meta-RCNN~\cite{DBLP:conf/iccv/YanCXWLL19}$_{ICCV19}$ 
        & 19.9 & 25.5 & 35 & 45.7 & 51.5
        & 10.4 & 19.4 & 29.6 & 34.8 & 45.4
        & 14.3 & 18.2 & 27.5 & 41.2 & 48.1\\
        & FsDetView~\cite{DBLP:conf/eccv/XiaoM20}$_{ECCV20}$  
        & 24.2 & 35.3 & 42.4 & 49.1 & 57.4 
        & 21.6 & 24.6 & 31.9 & 37.0 & 45.7
        & 31.2 & 30.0 & 37.2 & 43.8 & 49.6\\
        & CME~\cite{DBLP:conf/cvpr/LiYLLJY21}$_{CVPR21}$
        & 41.5 & 47.5 & 50.4 & 58.2 & 60.9 
        & 27.2 & 30.2 & \textcolor{blue}{41.4} & \textcolor{red}{42.5} & 46.8
        & 34.3 & 39.6 & {45.1} & 48.3 & 51.5\\
        & Wu et al.~\cite{DBLP:journals/corr/abs-2103-01077}$_{Arxiv21}$ & 43.8 & \textcolor{blue}{47.8} & 50.3 & 55.4 & 61.7
        & \textcolor{red}{31.2} & 30.5 & {41.2} & \textcolor{blue}{42.2} & \textcolor{blue}{48.3}
        & 35.5 & 39.7 & 43.9 & 50.6 & 53.5\\
        \midrule
        \multirow{2}{*}{\shortstack{Metric \\ Learning}} & RepMet~\cite{DBLP:conf/cvpr/KarlinskySHSAFG19}$_{CVPR19}$ & 26.1 & 32.9 & 34.4 & 38.6 & 41.3 
        & 17.2 & 22.1 & 23.4 & 28.3 & 35.8
        & 27.5 & 31.1 & 31.5 & 34.4 & 37.2\\
        & FSCE~\cite{DBLP:conf/cvpr/SunLCYZ21}$_{CVPR21}$ 
        & 44.2 & 43.8 & \textcolor{blue}{51.4} & \textcolor{red}{61.9} & \textcolor{red}{63.4}
        & 23.7 & 30.6 & 38.4 & 43.0 & \textcolor{red}{48.5}
        & 37.2 & {41.9} & \textcolor{red}{47.5} & \textcolor{blue}{54.6} & \textcolor{red}{58.5}\\
        \midrule
        \multirow{6}{*}{\shortstack{Transfer \\ Learning}} & MPSR~\cite{DBLP:conf/cvpr/SunLCYZ21}$_{ECCV20}$ 
        & 41.7 & 43.1 & {51.4} & {55.2} & {61.8}
        & 24.4 & 29.3 & 39.2 & 39.9 & {47.8}
        & 35.6 & {40.6} & {42.3} & {48.0} & {49.7}\\
        & FRCN-ft~\cite{DBLP:conf/icml/WangH0DY20}$_{ICML20}$ 
        & 8.2 & 20.3 & 29.0 & 40.1 & 45.5
        & 13.4 & 20.6 & 28.6 & 32.4 & 38.8
        & 19.6 & 20.8 & 28.7 & 42.2 & 42.1\\
        & TFA w/cos~\cite{DBLP:conf/icml/WangH0DY20}$_{ICML20}$ & 39.8 & 36.1 & 44.7 & 55.7 & 56.0
        & 23.5 & 26.9 & 34.1 & 35.1 & 39.1 
        & 30.8 & 34.8 & 42.8 & 49.5 & 49.8\\
        & Halluc~\cite{DBLP:conf/cvpr/ZhangW21a}$_{CVPR21}$ & \textcolor{red}{47.0} & 44.9 & 46.5 & 54.7 & 54.7 
        & 26.3 & \textcolor{blue}{31.8} & 37.4 & 37.4 & 41.2
        & \textcolor{blue}{40.4} & \textcolor{blue}{42.1} & 43.3 & 51.4 & 49.6\\
        & Fan et al.~\cite{DBLP:conf/cvpr/FanMLS21}$_{CVPR21}$ 
        & 42.4 & 45.8 & 45.9 & 53.7 & 56.1 
        & 21.7 & 27.8 & 35.2 & 37.0 & 40.3
        & 30.2 & 37.6 & 43.0 & 49.7 & 50.1\\
        \cmidrule{2-17}
        & \textbf{Our Method} &
        \textcolor{blue}{46.7} & \textcolor{red}{53.1} & \textcolor{red}{53.8} & \textcolor{blue}{61.0} & \textcolor{blue}{62.1}
        & \textcolor{blue}{30.1} & \textcolor{red}{34.2} & \textcolor{red}{41.6} & 41.9 & 44.8
        & \textcolor{red}{41.0} & \textcolor{red}{46.0} & \textcolor{blue}{47.2} & \textcolor{red}{55.4} & \textcolor{blue}{55.6}\\
        \bottomrule
    \end{tabular}
    \caption{Novel class AP50~(nAP50) results on PASCAL VOC. \textcolor{red}{Red} and \textcolor{blue}{Blue} indicate the best and the second best results respectively. Our proposed method significantly outperforms baseline TFA and achieves competitive results compared with other methods.}
    \label{table1}
\end{table*}


\begin{figure}[t]
    \centering
    \includegraphics[width=\columnwidth]{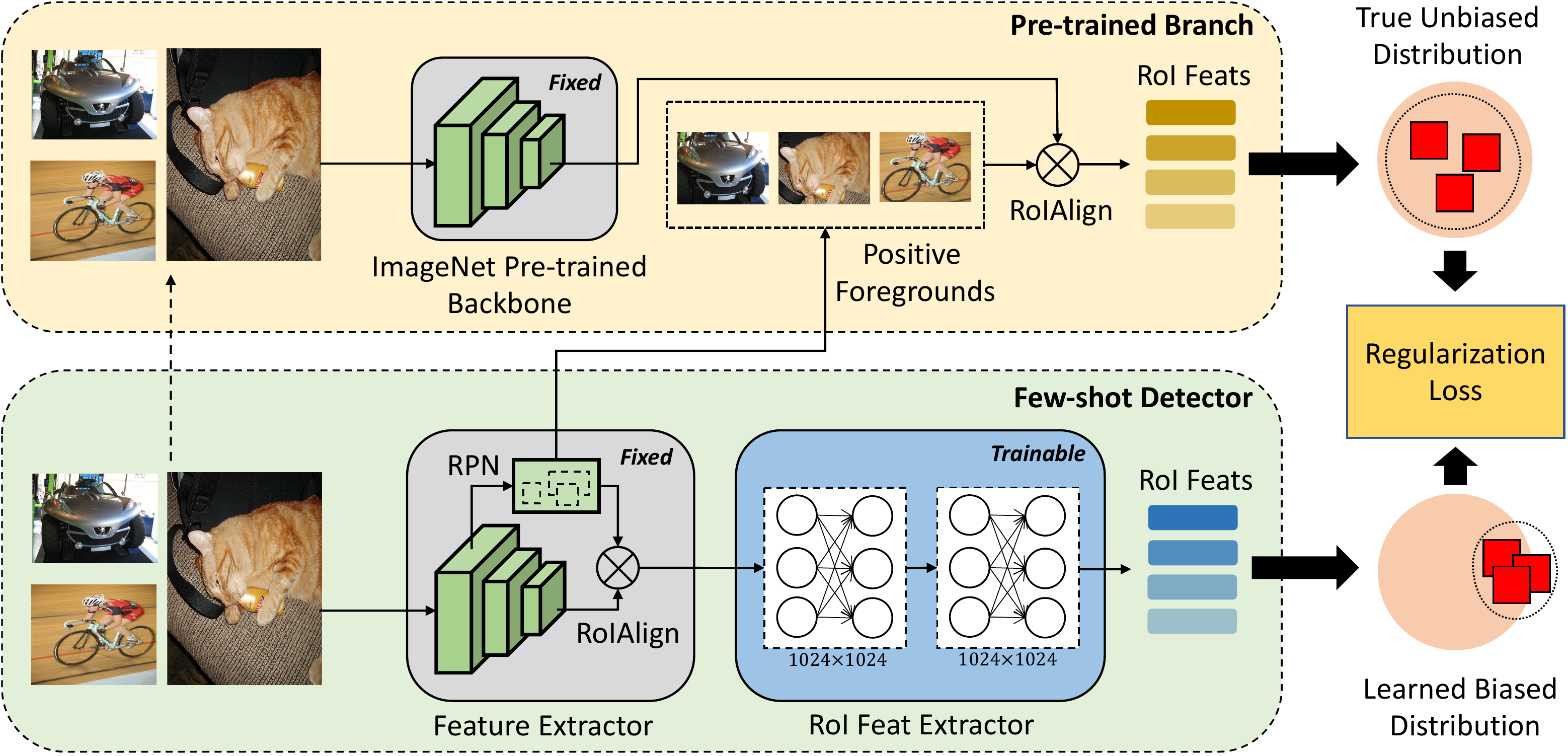}
    \caption{Illustration of the fine-tuning with distribution regularization. We use the ImageNet pre-trained network as a regularizer to guide the feature learning.}
    \label{fig3}
\end{figure}

\subsection{Distribution Regularization}

Through distribution calibration, the lack of inner-class variation issue is addressed. However, the primary challenge in the high-shot regime remains unsolved: the misalignment between the learned distributions and the true distributions. Commonly, the ImageNet pre-trained network is used to initialize the base detector. Considering that ImageNet pre-trained networks are rich in knowledge, they can provide useful guidance for feature learning during fine-tuning\cite{DBLP:conf/iccv/QiaoZLQWZ21}. Therefore, we propose to further fine-tune the detector with distribution regularization. 

Our proposed distribution regularization strategy is shown in Figure~\ref{fig3}. The ImageNet pre-trained network, which is used in parameter initializing in the base model training, is used for regularizing and guiding the fine-tuning. Specifically, let $X$ denote an image during fine-tuning. We feed $X$ and all positive foreground regions~(output by RPN in the few-shot detector) into the ImageNet pre-trained network to extract RoI features $\{ \hat{x_i} \}$. Let $\{ x_i \}$ denote the positive RoI features produced by the few-shot detector, we use a distance-based loss function to enforce the closeness between $\{ x_i \}$ and $\{ \hat{x_i} \}$ during fine-tuning:

\begin{equation}
    L_{reg} = \frac{1}{N} \sum_{i}^{N} ||x_i - \hat{x_i}||_2
\end{equation}

\noindent
where $N$ denotes the number of foreground regions. The joint loss is a combination of prediction loss and regularization loss:

\begin{equation}
    L = L_{cls} + L_{bbox} + \lambda L_{reg}
\end{equation}

\noindent
where $\lambda$ balances the detection loss and regularization loss. To make sure RoI features are learnable, the RoI feature extractor is not fixed during fine-tuning. Through our proposed distribution regularization, the semantic information contained in the ImageNet-pretrained network is fully utilized, and the biased learned distributions are implicitly aligned. 

\begin{table*}[t]
    \centering
    \begin{tabular}{l|ccc|ccc|ccc}
        \toprule
        \multirow{2}{*}{Methods} &  \multicolumn{3}{c|}{shot=1} & 
        \multicolumn{3}{c|}{shot=2} &
        \multicolumn{3}{c}{shot=3}\\
         & nAP & nAP$_{50}$ & nAP$_{75}$
         & nAP & nAP$_{50}$ & nAP$_{75}$
         & nAP & nAP$_{50}$ & nAP$_{75}$\\
         \midrule
         MPSR~\cite{DBLP:conf/cvpr/SunLCYZ21}$_{ECCV20}$ & 2.3 & 4.1 & 2.3
         & 3.5 & 6.3 & 3.4
         & 5.2 & 9.5 & 5.1\\
         FsDetView~\cite{DBLP:conf/eccv/XiaoM20}$_{ECCV20}$ & 3.2 & \textcolor{blue}{8.9} & 1.4
         & {4.9} & \textcolor{blue}{13.3} & 2.3
         & 6.7 & \textcolor{red}{18.6} & 2.9\\
         Hallucination~\cite{DBLP:conf/cvpr/ZhangW21a}$_{CVPR21}$ & \textcolor{blue}{4.4} & 7.5 & \textcolor{blue}{4.9}
         & \textcolor{blue}{5.6} & 9.9 & \textcolor{blue}{5.9}
         & \textcolor{blue}{7.2} & 13.3 & \textcolor{blue}{7.4}\\
         \midrule
         TFA w/cos~\cite{DBLP:conf/icml/WangH0DY20}$_{ICML20}$ & 3.4 & 5.8 & 3.8
         & 4.6 & 8.3 & 4.8
         & 6.6 & 12.1 & 6.5\\
         \textbf{TFA w/cos + DC~(ours)} & {5.1} & {9.4} & {5.0}
         & {6.0} & {11.4} & {6.0}
         & {7.3} & {14.5} & {7.0}\\
         \textbf{TFA w/cos + DC + DR~(ours)} & \textcolor{red}{5.7} & \textcolor{red}{10.6} & \textcolor{red}{5.8}
         & \textcolor{red}{7.1} & \textcolor{red}{13.3} & \textcolor{red}{7.0}
         & \textcolor{red}{8.6} & \textcolor{blue}{17.2} & \textcolor{red}{7.9}\\
         \bottomrule
    \end{tabular}
    \caption{Detection results of novel classes on COCO benchmark under low-shot settings~(shot=1/2/3). "DC" indicates distribution calibration and "DR" denotes distribution regularization. Our proposed method can achieve new state-of-the-art performance.}
    \label{table2}
\end{table*}

\section{Experiments}
We conduct extensive experiments on PASCAL VOC~(07+12) and MS COCO benchmarks. Our proposed method yields new state-of-the-art performance in most low-shot settings~(shot<5) and also achieves competitive results in high-shot settings~(shot$\ge$5). We employ the same setting~(base/novel split, base/novel samples) as in \cite{DBLP:conf/icml/WangH0DY20} for fair comparisons. Before presenting experimental results, we first elaborate on benchmark datasets and implementation details.

\subsection{Benchmark Datasets}
\textbf{PASCAL VOC.} For PASCAL VOC, all classes are divided into 15 base classes and 5 novel classes. VOC07 and 12 trainval-sets are used for base model training. Following TFA, the same three base/novel class splits~(namely split 1, 2, 3) and base/novel instances are used in experiments. We report AP50 of both base and novel classes in all settings.

\begin{table*}[t]
    \centering
    \begin{tabular}{c|l|cccccc|cccccc}
        \toprule
        \multirow{2}{*}{Type} & \multirow{2}{*}{Methods} & \multicolumn{6}{c|}{{shot=10}} & \multicolumn{6}{c}{{shot=30}}\\
        & & mAP & AP$_{50}$ & AP$_{75}$ & AP$_{s}$ & AP$_{m}$ & AP$_{l}$ & mAP & AP$_{50}$ & AP$_{75}$ & AP$_{s}$ & AP$_{m}$ & AP$_{l}$\\
        \midrule
        \multirow{4}{*}{\shortstack{Meta \\ Learning}} 
        & FSRW~\cite{DBLP:conf/iccv/KangLWYFD19}$_{ICCV19}$ & 5.6 & 12.3 & 4.6 & 0.9 & 3.5 & 10.5 & 9.1 & 19.0 & 7.6 & 0.8 & 4.9 & 16.8\\
        & Meta-Det~\cite{DBLP:conf/iccv/WangRH19}$_{ICCV19}$ & 7.1 & 14.6 & 6.1 & 1.0 & 4.1 & 12.2 & 11.3 & 21.7 & 8.1 & 1.1 & 6.2 & 17.3\\
        & Meta-RCNN~\cite{DBLP:conf/iccv/YanCXWLL19}$_{ICCV19}$ & 8.7 & 19.1 & 6.6 & 2.3 & 7.7 & 14.0 & 12.4 & 25.3 & 10.8 & 2.8 & 11.6 & 19.0\\
        & Wu et al.~\cite{DBLP:journals/corr/abs-2103-01077}$_{Arxiv21}$ & 11.0 & - & \textcolor{blue}{10.7} & \textcolor{blue}{4.5} & \textcolor{blue}{11.2} & \textcolor{blue}{17.3} & 15.6 & - & 15.7 & \textcolor{blue}{4.7} & \textcolor{blue}{15.1} & \textcolor{blue}{25.1}\\
        \midrule
        \shortstack{Metric \\ Learning}
        & FSCE~\cite{DBLP:conf/cvpr/SunLCYZ21}$_{CVPR21}$ & \textcolor{blue}{11.9} & - & 10.5 & - & - & - & \textcolor{blue}{16.4} & - & \textcolor{blue}{16.2} & - & - & -\\
        \midrule
        \multirow{6}{*}{\shortstack{Transfer \\ Learning}}
        & MPSR~\cite{DBLP:conf/eccv/WuL0W20}$_{ECCV20}$ & 9.8 & 17.9 & 9.7 & 3.3 & 9.2 & 16.1 & 14.1 & 25.4 & 14.2 & 4.0 & 12.9 & 23.0\\
        & Li et al.~\cite{DBLP:conf/cvpr/LiZCWT0VL21}$_{CVPR21}$ & 11.3 & \textcolor{blue}{20.3} & - & - & - & - & 15.1 & \textcolor{blue}{29.4} & - & - & - & -\\
        \cmidrule{2-14}
        & TFA w/cos~\cite{DBLP:conf/icml/WangH0DY20}$_{ICML20}$ & 10.0 & 19.1 & 9.3 & - & - & - & 13.7 & 24.7 & 13.2 & - & - & -\\
         & \textbf{TFA w/cos + DC~(ours)} & {9.9} & {19.0} & {9.0} & {4.3} & {8.9} & {15.9} & {13.3} & {25.1} & {13.1} & {5.6} & {12.3} & {20.5}\\
         & \textbf{TFA w/cos + DR~(ours)} & {12.1} & {23.9} & {11.1} & {4.0} & {11.7} & {19.7} & {16.7} & {31.2} & {16.2} & {4.0} & {17.3} & {25.1}\\
        & \textbf{TFA w/cos + DC + DR~(ours)} & \textcolor{red}{12.5} & \textcolor{red}{25.3} & \textcolor{red}{11.1} & \textcolor{red}{5.0} & \textcolor{red}{12.2} & \textcolor{red}{19.6} & \textcolor{red}{17.1} & \textcolor{red}{31.8} & \textcolor{red}{16.5} & \textcolor{red}{6.4} & \textcolor{red}{17.4} & \textcolor{red}{25.8}\\
        \bottomrule 
    \end{tabular}
    \caption{Performance comparisons on COCO benchmark. "DC" and "DR" denote distribution calibration and distribution regularization, respectively. Compared with existing state-of-the-art FSOD detectors, our method achieves competitive performance.}
    \label{table3}
\end{table*}

\begin{table*}[t]
    \centering
    \begin{tabular}{c|c|ccccc|ccccc|ccccc}
         \toprule 
        \multirow{2}{*}{} & 
        \multirow{2}{*}{Ablation} & 
         \multicolumn{5}{c|}{Split1} & \multicolumn{5}{c|}{Split2} & \multicolumn{5}{c}{Split3}\\
        & & 1 & 2 & 3 & 5 & 10
        & 1 & 2 & 3 & 5 & 10
        & 1 & 2 & 3 & 5 & 10\\
        \midrule
        \multirow{3}{*}{\shortstack{Novel \\ Classes}} & + DC (w/o SC) 
        & 44.6 & 50.9 & 51.3 & 58.2 & 58.0
        & 27.6 & 33.4 & 39.4 & 36.4 & 41.0
        & 37.3 & 40.7 & 42.2 & 51.0 & 51.2\\
        & + DC (w SC) 
        & 45.8 & 52.4 & 52.9 & 58.7 & 58.6
        & 28.2 & 33.8 & 40.5 & 37.9 & 42.3
        & 38.0 & 41.8 & 43.8 & 51.7 & 51.7\\
        & \cellcolor[gray]{.9}Improvement$\uparrow$ 
        & \cellcolor[gray]{.9}{+1.2} & \cellcolor[gray]{.9}{+1.5} & \cellcolor[gray]{.9}{+1.6} & \cellcolor[gray]{.9}{+0.5} & \cellcolor[gray]{.9}{+0.6}
        & \cellcolor[gray]{.9}{+0.6} & \cellcolor[gray]{.9}{+0.4} & \cellcolor[gray]{.9}{+0.9} & \cellcolor[gray]{.9}{+1.5} & \cellcolor[gray]{.9}{+1.3}
        & \cellcolor[gray]{.9}{+0.7} & \cellcolor[gray]{.9}{+1.1} & \cellcolor[gray]{.9}{+1.6} & \cellcolor[gray]{.9}{+0.7} & \cellcolor[gray]{.9}{+0.5}\\
        \midrule
        \multirow{4}{*}{\shortstack{Base \\ Classes}}
        & TFA w/cos~\cite{DBLP:conf/icml/WangH0DY20}
        & \textbf{79.6} & \textbf{78.9} & 79.1 & \textbf{79.3} & 78.4
        & \textbf{79.5} & \textbf{79.7} & 78.8 & 78.9 & 78.5
        & \textbf{80.3} & 79.9 & 80.4 & \textbf{80.2} & 79.9\\
        \cmidrule{2-17}
        & + DC (w/o SC) 
        & 78.6 & 78.1 & 78.8 & 78.6 & 79.0
        & 78.7 & 78.1 & 79.0 & 79.5 & 79.3
        & 79.4 & 79.7 & 80.3 & 79.7 & 80.0\\
        & + DC (w SC)  
        & 79.1 & 78.4 & \textbf{79.3} & 79.0 & \textbf{79.7}
        & 79.2 & 78.8 & \textbf{79.5} & \textbf{79.9} & \textbf{79.9}
        & 79.7 & \textbf{80.1} & \textbf{80.7} & {80.0} & \textbf{80.4}\\
        & \cellcolor[gray]{.9}Improvement$\uparrow$ 
        & \cellcolor[gray]{.9}{+0.5} & \cellcolor[gray]{.9}{+0.3} & \cellcolor[gray]{.9}{+0.5} & \cellcolor[gray]{.9}{+0.4} & \cellcolor[gray]{.9}{+0.7}
        & \cellcolor[gray]{.9}{+0.5} & \cellcolor[gray]{.9}{+0.7} & \cellcolor[gray]{.9}{+0.5} & \cellcolor[gray]{.9}{+0.4} & \cellcolor[gray]{.9}{0.6}
        & \cellcolor[gray]{.9}{+0.3} & \cellcolor[gray]{.9}{+0.4} & \cellcolor[gray]{.9}{+0.4} & \cellcolor[gray]{.9}{+0.3} & \cellcolor[gray]{.9}{+0.4}\\
        \bottomrule
    \end{tabular}
    \caption{Ablation experiments on shift compensation. "DC" and "SC" denote distribution calibration and shift compensation, respectively. Our proposed shift compensation can boost performance of distribution calibration on both novel and base classes. }
    \label{table4}
\end{table*}

\noindent
\textbf{COCO.} On COCO benchmark, 80 categories are divided into 60 base classes and 20 novel classes. Evaluations are performed under shot=10/30 settings on COCO 2017 val-set. To further validate the effectiveness of our proposed method, we conduct experiments in shot=1/2/3 settings. Experimental setup is the same as in previous works~\cite{DBLP:conf/icml/WangH0DY20,DBLP:conf/cvpr/ZhangW21a}. COCO-style metrics are reported.


\subsection{Implementation Details}
For base model training on PASCAL VOC, we train model 50k iterations on 4 GPUs with base-size 4. The initial learning rate is set to 0.02 and reduced by 0.1 at 40k iterations. SGD optimizer with momentum 0.9 and weight decay 1e-4 is used. 
Positive RoI features~(IOU$\ge$0.5 with ground truth) are used to obtain the distribution of each base class. Same to TFA, we also use Faster-RCNN as the detection model and ImageNet pre-trained ResNet-101 and FPN as the feature extractor. For few-shot fine-tuning, the initial learning rate is set to 0.002, and fine-tuning steps are adjusted according to the number of shots. Model will be fine-tuned on a balanced set consisting of data from both base and novel classes where each class contains $K$ samples, namely $K$ shot setting. We firstly fine-tune the detector with distribution calibration and then we fine-tune the detector with distribution regularization. To maintain a balanced base-novel set during fine-tuning, we also sample $M$ base class synthetic samples. Implementation details on COCO benchmark are similar.

As for hyper-parameters, $\alpha$ in Equation~\ref{e6} is experimentally set to $10^{-4}$ or $10^{-3}$. Numbers of new generated sample $M$ is set to 1. $\lambda$ in regularization-based fine-tuning is set to $10^{-4}$. $\theta$ in shift compensation is set to 0.9999. As for $k$ in distribution calibration, we give more details in ablation study.


\subsection{Main Results}
\textbf{PASCAL VOC.} Detection results of our proposed method are shown in Table~\ref{table1}. Our proposed method yields significant performance gain in both low-shot settings~(shot<5) and high-shot settings~(shot$\ge$5) compared to baseline TFA. Meanwhile, compared with other FSOD methods, our proposed distribution calibration method can achieve new state-of-the-art performance in most low-shot settings~(shot<5). In high-shot settings, our method can also achieve comparable results. More importantly, our proposed method can achieve more stable results: compared with other FSOD methods whose performance are unstable across different shots, our method can achieve competitive performance in not only low-shot but also high-shot settings.



\noindent
\textbf{COCO.} Experimental results on COCO benchmark under low-shot settings are shown in Table~\ref{table2}. Our method can largely outperform baseline TFA. Compared with existing methods, the proposed method achieves new state-of-the-art performance. Experimental results on COCO benchmark under high-shot settings are shown in Table~\ref{table3}. Our method can achieve competitive results compared with existing FSOD methods. Compared with existing transfer learning based FSOD methods, our method yields significant improvement. Experiments in COCO low-shot and high-shot settings further validate the effectiveness of our proposed method. Meanwhile, we test the effect of DC and DR, respectively. 
Results show that when DC and DR are combined together, the performance comes to the best, which demonstrates that DC and DR are not mutually exclusive but can reinforce each other.


\begin{table*}[t]
    \centering
    \begin{tabular}{c|c|ccccc|ccccc|cccccc}
         \toprule 
        \multirow{2}{*}{} &
        \multirow{2}{*}{Ablation} & 
         \multicolumn{5}{c|}{Split1} & \multicolumn{5}{c|}{Split2} & \multicolumn{5}{c}{Split3}\\
         & & 1 & 2 & 3 & 5 & 10
        & 1 & 2 & 3 & 5 & 10
        & 1 & 2 & 3 & 5 & 10\\
        \midrule
        \multirow{5}{*}{\shortstack{Novel \\ Classes}} 
        & TFA w/cos~\cite{DBLP:conf/icml/WangH0DY20}
        & 39.8 & 36.1 & 44.7 & 55.7 & 56.0
        & 23.5 & 26.9 & 34.1 & 35.1 & 39.1
        & 30.8 & 34.8 & 42.8 & 49.5 & 49.8\\
        \cmidrule(r){2-17}
        & + DC 
        & 45.8 & 52.4 & 52.9 & 58.7 & 58.6
        & 28.2 & 33.8 & 40.5 & 37.9 & 42.3
        & 38.0 & 41.8 & 43.8 & 51.7 & 51.7\\
        & \cellcolor[gray]{.9}Improvement$\uparrow$ 
        & \cellcolor[gray]{.9}+6.0 & \cellcolor[gray]{.9}+16.3 & \cellcolor[gray]{.9}+8.2 & \cellcolor[gray]{.9}+3.0 & \cellcolor[gray]{.9}+2.6
        & \cellcolor[gray]{.9}+4.7 & \cellcolor[gray]{.9}+6.9 & \cellcolor[gray]{.9}+6.5 & \cellcolor[gray]{.9}+2.8 & \cellcolor[gray]{.9}+3.2
        & \cellcolor[gray]{.9}+7.2 & \cellcolor[gray]{.9}+7.0 & \cellcolor[gray]{.9}+1.0 & \cellcolor[gray]{.9}+2.2 & \cellcolor[gray]{.9}+1.9\\
        
        \cmidrule(r){2-17}
        & + DC + DR
        & 46.7 & 53.1 & 53.8 & 61.0 & 62.1
        & 30.1 & 34.2 & 41.6 & 41.9 & 44.8
        & 41.0 & 46.0 & 47.2 & 55.4 & 56.6\\
        & \cellcolor[gray]{.9}Improvement$\uparrow$ 
        & \cellcolor[gray]{.9}{+0.9} & \cellcolor[gray]{.9}{+0.7} & \cellcolor[gray]{.9}{+0.9} & \cellcolor[gray]{.9}{+2.3} & \cellcolor[gray]{.9}{+3.5}
        & \cellcolor[gray]{.9}{+1.9} & \cellcolor[gray]{.9}{+0.4} & \cellcolor[gray]{.9}{+1.1} & \cellcolor[gray]{.9}{+4.0} & \cellcolor[gray]{.9}{+2.5}
        & \cellcolor[gray]{.9}{+3.0} & \cellcolor[gray]{.9}{+4.2} & \cellcolor[gray]{.9}{+3.4} & \cellcolor[gray]{.9}{+3.7} & \cellcolor[gray]{.9}{+4.9}\\
        \midrule
        \multirow{5}{*}{\shortstack{Base \\ Classes}}
        & TFA w/cos~\cite{DBLP:conf/icml/WangH0DY20}
        & \textbf{79.6} & {78.9} & 79.1 & {79.3} & 78.4
        & \textbf{79.5} & {79.7} & 78.8 & 78.9 & 78.5
        & {80.3} & \textbf{79.9} & \textbf{80.4} & \textbf{80.2} & 79.9\\
        \cmidrule(r){2-17}
        & + DC
        & 79.1 & 78.4 & 79.3 & 79.0 & 79.7
        & 79.2 & 78.8 & 79.5 & 79.9 & 79.9
        & 79.7 & 80.1 & 80.7 & 80.0 & 80.4\\
        & \cellcolor[gray]{.9}Improvement$\uparrow$ 
        & \cellcolor[gray]{.9}{-0.5} & \cellcolor[gray]{.9}{-0.5} & \cellcolor[gray]{.9}{+0.2} & \cellcolor[gray]{.9}{-0.3} & \cellcolor[gray]{.9}{+1.3}
        & \cellcolor[gray]{.9}{-0.3} & \cellcolor[gray]{.9}{-0.9} & \cellcolor[gray]{.9}{+0.7} & \cellcolor[gray]{.9}{+1.0} & \cellcolor[gray]{.9}{+1.4}
        & \cellcolor[gray]{.9}{-0.6} & \cellcolor[gray]{.9}{+0.2} & \cellcolor[gray]{.9}{+0.3} & \cellcolor[gray]{.9}{+0.0} & \cellcolor[gray]{.9}{+0.5}\\
        \cmidrule(r){2-17}
        & + DC + DR
        & 79.2 & \textbf{79.2} & \textbf{79.7} & \textbf{79.3} & \textbf{79.6}
        & 79.3 & \textbf{80.0} & \textbf{80.3} & \textbf{80.2} & \textbf{80.3}
        & \textbf{80.3} & {79.8} & {79.9} & {80.0} & \textbf{80.9}\\
        & \cellcolor[gray]{.9}Improvement$\uparrow$ 
        & \cellcolor[gray]{.9}{+0.1} & \cellcolor[gray]{.9}{+0.8} & \cellcolor[gray]{.9}{+0.4} & \cellcolor[gray]{.9}{+0.3} & \cellcolor[gray]{.9}{-0.1}
        & \cellcolor[gray]{.9}{+0.1} & \cellcolor[gray]{.9}{+1.2} & \cellcolor[gray]{.9}{+0.8} & \cellcolor[gray]{.9}{+0.3} & \cellcolor[gray]{.9}{0.4}
        & \cellcolor[gray]{.9}{+0.6} & \cellcolor[gray]{.9}{-0.3} & \cellcolor[gray]{.9}{-0.8} & \cellcolor[gray]{.9}{+0.0} & \cellcolor[gray]{.9}{+0.5}\\
        \bottomrule
    \end{tabular}
    \caption{Ablation experiments of the distribution calibration and distribution regularization on PASCAL VOC. "DC" indicates distribution calibration and "DR" denotes distribution regularization.}
    \label{table5}
\end{table*}

\subsection{Ablation Studies}

\textbf{Ablation for shift compensation.}
Ablation studies of our proposed shift compensation on PASCAL VOC benchmark is shown in Table~\ref{table4}. "w" and "w/o" indicates distribution calibration with and without shift compensation, respectively. Results show that SC is helpful in boosting performance of DC for both novel classes and base classes. For base classes, in few cases, there is a slight performance degradation compared with baseline TFA. Fortunately, this performance gap can be compensated by our proposed SC in a certain degree.

\noindent
\textbf{Ablation for distribution calibration and distribution regularization.}
We conduct ablation studies on distribution calibration and distribution regularization on PASCAL VOC benchmark. Results are shown in Table~\ref{table5}. We can see that in low-shot settings the improvements mainly come from distribution calibration, while in high-shot settings the improvements led by distribution regularization is higher. This validates our assumption that the primary issues for low-shot and high-shot regimes are different from each other. Besides, for base classes, detection performance is slightly compromised in only few cases, our proposed method can achieve better performance compared with baseline TFA in most cases.

\begin{figure}[t]
    \centering
    \includegraphics[width=\columnwidth]{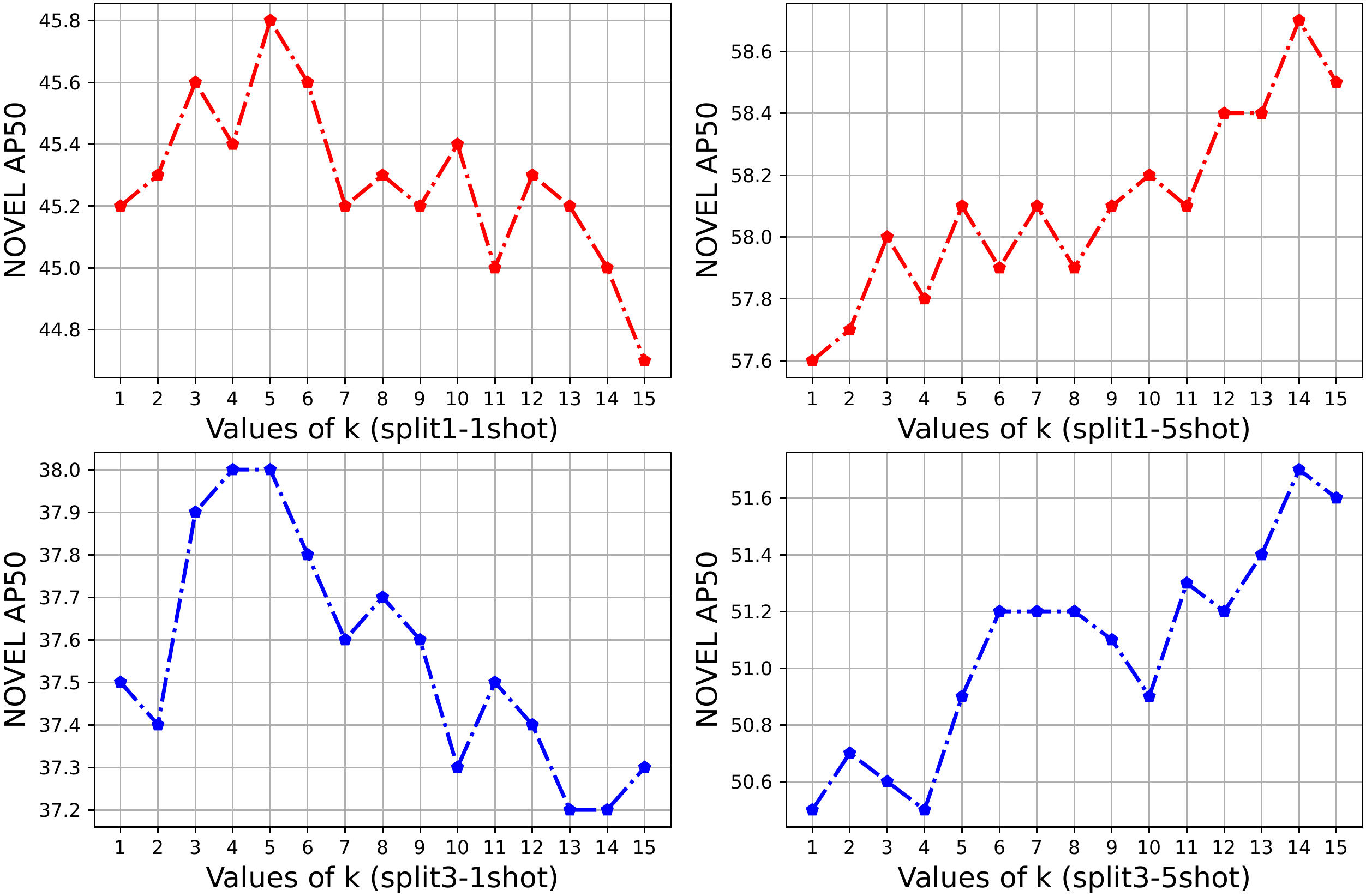}
    \caption{The effect of different $k$ in distribution calibration.}
    \label{fig5}
\end{figure}

\begin{figure}[t]
    \centering
    \includegraphics[width=1.0\columnwidth]{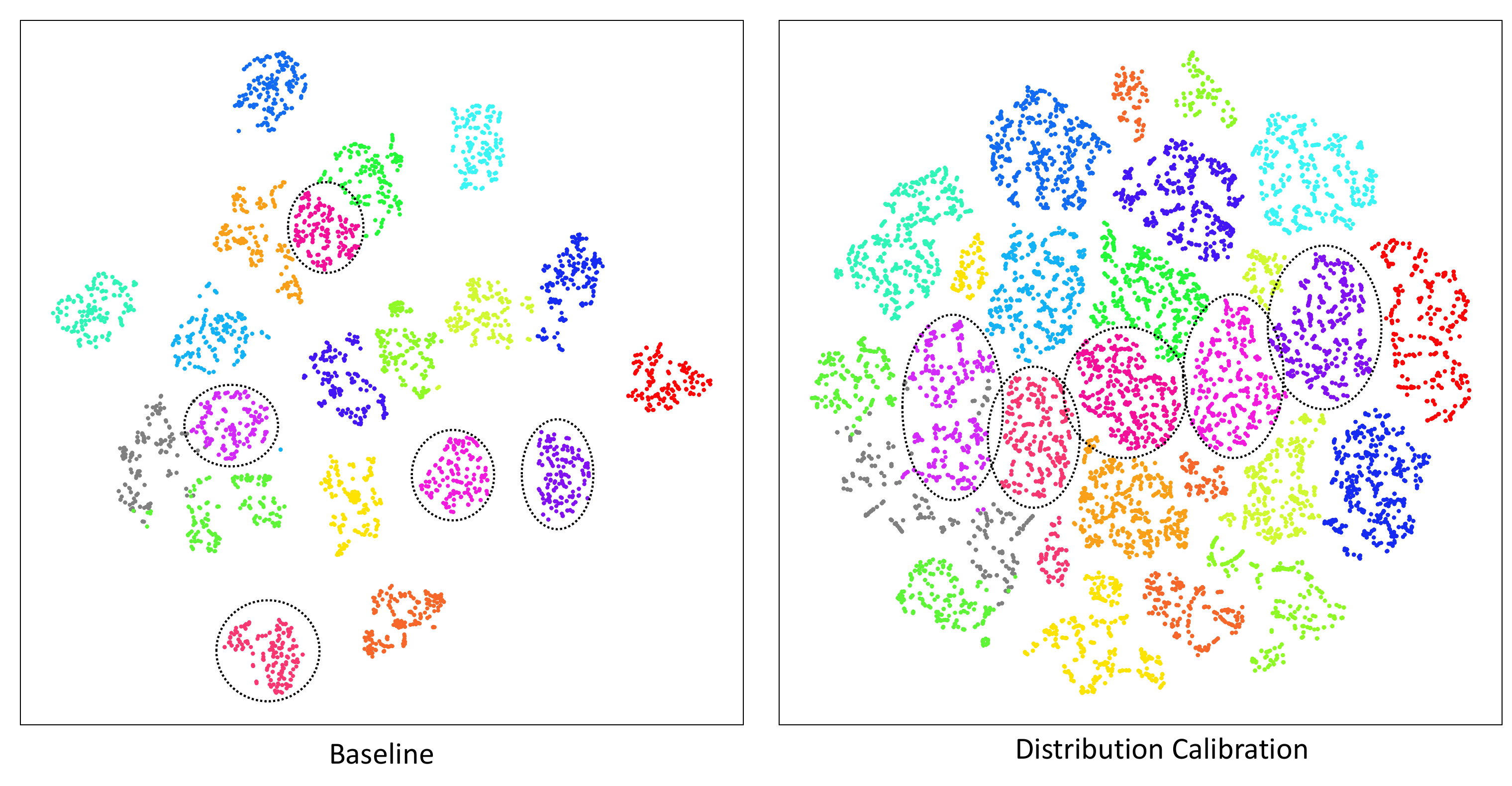}
    \caption{t-SNE visualization of synthetic training samples generated by distribution calibration. 
    }
    \label{fig6}
\end{figure}

\noindent
\textbf{Ablation for $k$ in Distribution Calibration.}
In the second stage distribution calibration, $k$ controls the number of base classes that are used to calibrate the novel distribution. 
For investigation, we test the effects of different $k$, and results are shown in Figure~\ref{fig5}. We find that in low-shot settings~(shot<5), smaller $k$~(normally between 3-5) leads to the best performance. Higher $k$ might damage the performance. In high-shot settings~(shot$\ge$5), higher $k$~(around 15 on PASCAL VOC) leads to the best result. 


\noindent
\textbf{Visualization Analysis.}
Figure~\ref{fig6} gives a t-SNE analysis of distribution calibration in split1 1-shot setting on PASCAL VOC. The left sub-figure shows the biased distribution in baseline TFA and the right sub-figure shows the calibrated distribution.
It can be seen that the inner-class variation of novel classes is enriched by generated samples. It is worth noting that Figure~\ref{fig6} is to show the distributions of the generated samples and calibrated distributions~\cite{DBLP:conf/nips/GaoSZZC18,DBLP:conf/cvpr/WangGHH18}. 


\section{Conclusions}
In this paper, we reveal the fundamental cause of shot-unstable in FSOD: the problem nature of few-shot detection is different in low-shot and high-shot cases. In the low-shot regime the primary challenge is the lack of inner-class variation, while in the high-shot regime the primary issue is a misalignment problem. Meanwhile, we propose to address these distinct two issues by transferring the learned knowledge more effectively. Our proposed method achieves competitive and more stable results.



\begin{acks}
This work was supported by National Natural Science Foundation of China (Grant No. 62176017 and No.41871283).
\end{acks}

\clearpage

\balance

\bibliographystyle{ACM-Reference-Format}
\bibliography{sample-sigconf-final}


\begin{thebibliography}{62}


\ifx \showCODEN    \undefined \def \showCODEN     #1{\unskip}     \fi
\ifx \showDOI      \undefined \def \showDOI       #1{#1}\fi
\ifx \showISBNx    \undefined \def \showISBNx     #1{\unskip}     \fi
\ifx \showISBNxiii \undefined \def \showISBNxiii  #1{\unskip}     \fi
\ifx \showISSN     \undefined \def \showISSN      #1{\unskip}     \fi
\ifx \showLCCN     \undefined \def \showLCCN      #1{\unskip}     \fi
\ifx \shownote     \undefined \def \shownote      #1{#1}          \fi
\ifx \showarticletitle \undefined \def \showarticletitle #1{#1}   \fi
\ifx \showURL      \undefined \def \showURL       {\relax}        \fi
\providecommand\bibfield[2]{#2}
\providecommand\bibinfo[2]{#2}
\providecommand\natexlab[1]{#1}
\providecommand\showeprint[2][]{arXiv:#2}

\bibitem[Cai and Vasconcelos(2017)]%
        {DBLP:journals/corr/abs-1712-00726}
\bibfield{author}{\bibinfo{person}{Zhaowei Cai} {and} \bibinfo{person}{Nuno
  Vasconcelos}.} \bibinfo{year}{2017}\natexlab{}.
\newblock \showarticletitle{Cascade {R-CNN:} Delving into High Quality Object
  Detection}.
\newblock \bibinfo{journal}{\emph{CoRR}}  \bibinfo{volume}{abs/1712.00726}
  (\bibinfo{year}{2017}).
\newblock


\bibitem[Chen et~al\mbox{.}(2018)]%
        {DBLP:conf/aaai/ChenWW018}
\bibfield{author}{\bibinfo{person}{Hao Chen}, \bibinfo{person}{Yali Wang},
  \bibinfo{person}{Guoyou Wang}, {and} \bibinfo{person}{Yu Qiao}.}
  \bibinfo{year}{2018}\natexlab{}.
\newblock \showarticletitle{{LSTD:} {A} Low-Shot Transfer Detector for Object
  Detection}. In \bibinfo{booktitle}{\emph{{AAAI}}}.
  \bibinfo{pages}{2836--2843}.
\newblock


\bibitem[Donahue et~al\mbox{.}(2014)]%
        {DBLP:conf/icml/DonahueJVHZTD14}
\bibfield{author}{\bibinfo{person}{Jeff Donahue}, \bibinfo{person}{Yangqing
  Jia}, \bibinfo{person}{Oriol Vinyals}, \bibinfo{person}{Judy Hoffman},
  \bibinfo{person}{Ning Zhang}, \bibinfo{person}{Eric Tzeng}, {and}
  \bibinfo{person}{Trevor Darrell}.} \bibinfo{year}{2014}\natexlab{}.
\newblock \showarticletitle{DeCAF: {A} Deep Convolutional Activation Feature
  for Generic Visual Recognition}. In \bibinfo{booktitle}{\emph{{ICML}}}.
  \bibinfo{pages}{647--655}.
\newblock


\bibitem[Dong et~al\mbox{.}(2021)]%
        {DBLP:conf/mm/DongYXHY21}
\bibfield{author}{\bibinfo{person}{Kaiqi Dong}, \bibinfo{person}{Wei Yang},
  \bibinfo{person}{Zhenbo Xu}, \bibinfo{person}{Liusheng Huang}, {and}
  \bibinfo{person}{Zhidong Yu}.} \bibinfo{year}{2021}\natexlab{}.
\newblock \showarticletitle{ABPNet: Adaptive Background Modeling for
  Generalized Few Shot Segmentation}. In \bibinfo{booktitle}{\emph{{ACM}
  Multimedia}}. \bibinfo{publisher}{{ACM}}, \bibinfo{pages}{2271--2280}.
\newblock


\bibitem[Dong et~al\mbox{.}(2018)]%
        {DBLP:conf/mm/DongZZYW18}
\bibfield{author}{\bibinfo{person}{Xuanyi Dong}, \bibinfo{person}{Linchao Zhu},
  \bibinfo{person}{De Zhang}, \bibinfo{person}{Yi Yang}, {and}
  \bibinfo{person}{Fei Wu}.} \bibinfo{year}{2018}\natexlab{}.
\newblock \showarticletitle{Fast Parameter Adaptation for Few-shot Image
  Captioning and Visual Question Answering}. In \bibinfo{booktitle}{\emph{{ACM}
  Multimedia}}. \bibinfo{publisher}{{ACM}}, \bibinfo{pages}{54--62}.
\newblock


\bibitem[Fan et~al\mbox{.}(2021)]%
        {DBLP:conf/cvpr/FanMLS21}
\bibfield{author}{\bibinfo{person}{Zhibo Fan}, \bibinfo{person}{Yuchen Ma},
  \bibinfo{person}{Zeming Li}, {and} \bibinfo{person}{Jian Sun}.}
  \bibinfo{year}{2021}\natexlab{}.
\newblock \showarticletitle{Generalized Few-Shot Object Detection without
  Forgetting}. In \bibinfo{booktitle}{\emph{{CVPR}}}.
  \bibinfo{pages}{4527--4536}.
\newblock


\bibitem[Finn et~al\mbox{.}(2017)]%
        {DBLP:conf/icml/FinnAL17}
\bibfield{author}{\bibinfo{person}{Chelsea Finn}, \bibinfo{person}{Pieter
  Abbeel}, {and} \bibinfo{person}{Sergey Levine}.}
  \bibinfo{year}{2017}\natexlab{}.
\newblock \showarticletitle{Model-Agnostic Meta-Learning for Fast Adaptation of
  Deep Networks}. In \bibinfo{booktitle}{\emph{{ICML}}}.
  \bibinfo{pages}{1126--1135}.
\newblock


\bibitem[Gao et~al\mbox{.}(2018)]%
        {DBLP:conf/nips/GaoSZZC18}
\bibfield{author}{\bibinfo{person}{Hang Gao}, \bibinfo{person}{Zheng Shou},
  \bibinfo{person}{Alireza Zareian}, \bibinfo{person}{Hanwang Zhang}, {and}
  \bibinfo{person}{Shih{-}Fu Chang}.} \bibinfo{year}{2018}\natexlab{}.
\newblock \showarticletitle{Low-shot Learning via Covariance-Preserving
  Adversarial Augmentation Networks}. In \bibinfo{booktitle}{\emph{NeurIPS}}.
  \bibinfo{pages}{983--993}.
\newblock


\bibitem[Hoyer et~al\mbox{.}(2021)]%
        {DBLP:journals/corr/abs-2111-14887}
\bibfield{author}{\bibinfo{person}{Lukas Hoyer}, \bibinfo{person}{Dengxin Dai},
  {and} \bibinfo{person}{Luc~Van Gool}.} \bibinfo{year}{2021}\natexlab{}.
\newblock \showarticletitle{DAFormer: Improving Network Architectures and
  Training Strategies for Domain-Adaptive Semantic Segmentation}.
\newblock \bibinfo{journal}{\emph{CoRR}}  \bibinfo{volume}{abs/2111.14887}
  (\bibinfo{year}{2021}).
\newblock


\bibitem[Johnson(2018)]%
        {DBLP:journals/corr/abs-1805-00500}
\bibfield{author}{\bibinfo{person}{Jeremiah~W. Johnson}.}
  \bibinfo{year}{2018}\natexlab{}.
\newblock \showarticletitle{Adapting Mask-RCNN for Automatic Nucleus
  Segmentation}.
\newblock \bibinfo{journal}{\emph{CoRR}}  \bibinfo{volume}{abs/1805.00500}
  (\bibinfo{year}{2018}).
\newblock


\bibitem[Kang et~al\mbox{.}(2019)]%
        {DBLP:conf/iccv/KangLWYFD19}
\bibfield{author}{\bibinfo{person}{Bingyi Kang}, \bibinfo{person}{Zhuang Liu},
  \bibinfo{person}{Xin Wang}, \bibinfo{person}{Fisher Yu},
  \bibinfo{person}{Jiashi Feng}, {and} \bibinfo{person}{Trevor Darrell}.}
  \bibinfo{year}{2019}\natexlab{}.
\newblock \showarticletitle{Few-Shot Object Detection via Feature Reweighting}.
  In \bibinfo{booktitle}{\emph{{ICCV}}}. \bibinfo{pages}{8419--8428}.
\newblock


\bibitem[Karlinsky et~al\mbox{.}(2019)]%
        {DBLP:conf/cvpr/KarlinskySHSAFG19}
\bibfield{author}{\bibinfo{person}{Leonid Karlinsky}, \bibinfo{person}{Joseph
  Shtok}, \bibinfo{person}{Sivan Harary}, \bibinfo{person}{Eli Schwartz},
  \bibinfo{person}{Amit Aides}, \bibinfo{person}{Rog{\'{e}}rio~Schmidt Feris},
  \bibinfo{person}{Raja Giryes}, {and} \bibinfo{person}{Alexander~M.
  Bronstein}.} \bibinfo{year}{2019}\natexlab{}.
\newblock \showarticletitle{RepMet: Representative-Based Metric Learning for
  Classification and Few-Shot Object Detection}. In
  \bibinfo{booktitle}{\emph{{CVPR}}}. \bibinfo{pages}{5197--5206}.
\newblock


\bibitem[Koch et~al\mbox{.}(2015)]%
        {koch2015siamese}
\bibfield{author}{\bibinfo{person}{Gregory Koch}, \bibinfo{person}{Richard
  Zemel}, \bibinfo{person}{Ruslan Salakhutdinov}, {et~al\mbox{.}}}
  \bibinfo{year}{2015}\natexlab{}.
\newblock \showarticletitle{Siamese Neural Networks for One-shot Image
  Recognition}. In \bibinfo{booktitle}{\emph{ICML}}. \bibinfo{pages}{0}.
\newblock


\bibitem[Kull et~al\mbox{.}(2017)]%
        {DBLP:conf/aistats/KullFF17}
\bibfield{author}{\bibinfo{person}{Meelis Kull}, \bibinfo{person}{Telmo
  de~Menezes~e Silva~Filho}, {and} \bibinfo{person}{Peter~A. Flach}.}
  \bibinfo{year}{2017}\natexlab{}.
\newblock \showarticletitle{Beta Calibration: A Well-founded and Easily
  Implemented Improvement on Logistic Calibration for Binary Classifiers}. In
  \bibinfo{booktitle}{\emph{{AISTATS}}}. \bibinfo{pages}{623--631}.
\newblock


\bibitem[Law and Deng(2018)]%
        {DBLP:conf/eccv/LawD18}
\bibfield{author}{\bibinfo{person}{Hei Law} {and} \bibinfo{person}{Jia Deng}.}
  \bibinfo{year}{2018}\natexlab{}.
\newblock \showarticletitle{CornerNet: Detecting Objects as Paired Keypoints}.
  In \bibinfo{booktitle}{\emph{{ECCV}}}. \bibinfo{pages}{765--781}.
\newblock


\bibitem[Lee et~al\mbox{.}(2020)]%
        {DBLP:conf/iclr/LeeCK20}
\bibfield{author}{\bibinfo{person}{Cheolhyoung Lee}, \bibinfo{person}{Kyunghyun
  Cho}, {and} \bibinfo{person}{Wanmo Kang}.} \bibinfo{year}{2020}\natexlab{}.
\newblock \showarticletitle{Mixout: Effective Regularization to Finetune
  Large-scale Pretrained Language Models}. In
  \bibinfo{booktitle}{\emph{{ICLR}}}.
\newblock


\bibitem[Lee et~al\mbox{.}(2019)]%
        {DBLP:conf/cvpr/LeeMRS19}
\bibfield{author}{\bibinfo{person}{Kwonjoon Lee}, \bibinfo{person}{Subhransu
  Maji}, \bibinfo{person}{Avinash Ravichandran}, {and} \bibinfo{person}{Stefano
  Soatto}.} \bibinfo{year}{2019}\natexlab{}.
\newblock \showarticletitle{Meta-Learning With Differentiable Convex
  Optimization}. In \bibinfo{booktitle}{\emph{{CVPR}}}.
  \bibinfo{pages}{10657--10665}.
\newblock


\bibitem[Li et~al\mbox{.}(2021a)]%
        {DBLP:conf/cvpr/LiYLLJY21}
\bibfield{author}{\bibinfo{person}{Bohao Li}, \bibinfo{person}{Boyu Yang},
  \bibinfo{person}{Chang Liu}, \bibinfo{person}{Feng Liu},
  \bibinfo{person}{Rongrong Ji}, {and} \bibinfo{person}{Qixiang Ye}.}
  \bibinfo{year}{2021}\natexlab{a}.
\newblock \showarticletitle{Beyond Max-Margin: Class Margin Equilibrium for
  Few-Shot Object Detection}. In \bibinfo{booktitle}{\emph{{CVPR}}}.
  \bibinfo{pages}{7363--7372}.
\newblock


\bibitem[Li et~al\mbox{.}(2021b)]%
        {DBLP:conf/cvpr/LiZCWT0VL21}
\bibfield{author}{\bibinfo{person}{Yiting Li}, \bibinfo{person}{Haiyue Zhu},
  \bibinfo{person}{Yu Cheng}, \bibinfo{person}{Wenxin Wang},
  \bibinfo{person}{Chek~Sing Teo}, \bibinfo{person}{Cheng Xiang},
  \bibinfo{person}{Prahlad Vadakkepat}, {and} \bibinfo{person}{Tong~Heng Lee}.}
  \bibinfo{year}{2021}\natexlab{b}.
\newblock \showarticletitle{Few-Shot Object Detection via Classification
  Refinement and Distractor Retreatment}. In
  \bibinfo{booktitle}{\emph{{CVPR}}}. \bibinfo{pages}{15395--15403}.
\newblock


\bibitem[Lin et~al\mbox{.}(2017)]%
        {DBLP:conf/iccv/LinGGHD17}
\bibfield{author}{\bibinfo{person}{Tsung{-}Yi Lin}, \bibinfo{person}{Priya
  Goyal}, \bibinfo{person}{Ross~B. Girshick}, \bibinfo{person}{Kaiming He},
  {and} \bibinfo{person}{Piotr Doll{\'{a}}r}.} \bibinfo{year}{2017}\natexlab{}.
\newblock \showarticletitle{Focal Loss for Dense Object Detection}. In
  \bibinfo{booktitle}{\emph{{ICCV}}}. \bibinfo{pages}{2999--3007}.
\newblock


\bibitem[Liu et~al\mbox{.}(2021)]%
        {DBLP:conf/mm/LiuMZYL21}
\bibfield{author}{\bibinfo{person}{Longyao Liu}, \bibinfo{person}{Bo Ma},
  \bibinfo{person}{Yulin Zhang}, \bibinfo{person}{Xin Yi}, {and}
  \bibinfo{person}{Haozhi Li}.} \bibinfo{year}{2021}\natexlab{}.
\newblock \showarticletitle{AFD-Net: Adaptive Fully-Dual Network for Few-shot
  Object Detection}. In \bibinfo{booktitle}{\emph{{ACM} Multimedia}}.
  \bibinfo{publisher}{{ACM}}, \bibinfo{pages}{2549--2557}.
\newblock


\bibitem[Liu et~al\mbox{.}(2016)]%
        {DBLP:conf/eccv/LiuAESRFB16}
\bibfield{author}{\bibinfo{person}{Wei Liu}, \bibinfo{person}{Dragomir
  Anguelov}, \bibinfo{person}{Dumitru Erhan}, \bibinfo{person}{Christian
  Szegedy}, \bibinfo{person}{Scott~E. Reed}, \bibinfo{person}{Cheng{-}Yang Fu},
  {and} \bibinfo{person}{Alexander~C. Berg}.} \bibinfo{year}{2016}\natexlab{}.
\newblock \showarticletitle{{SSD:} Single Shot MultiBox Detector}. In
  \bibinfo{booktitle}{\emph{{ECCV}}}. \bibinfo{pages}{21--37}.
\newblock


\bibitem[Oreshkin et~al\mbox{.}(2018)]%
        {DBLP:conf/nips/OreshkinLL18}
\bibfield{author}{\bibinfo{person}{Boris~N. Oreshkin},
  \bibinfo{person}{Pau~Rodr{\'{\i}}guez L{\'{o}}pez}, {and}
  \bibinfo{person}{Alexandre Lacoste}.} \bibinfo{year}{2018}\natexlab{}.
\newblock \showarticletitle{{TADAM:} Task dependent adaptive metric for
  improved few-shot learning}. In \bibinfo{booktitle}{\emph{NIPS}}.
  \bibinfo{pages}{719--729}.
\newblock


\bibitem[Pang et~al\mbox{.}(2019)]%
        {DBLP:conf/cvpr/PangCSFOL19}
\bibfield{author}{\bibinfo{person}{J. Pang}, \bibinfo{person}{K. Chen},
  \bibinfo{person}{J. Shi}, {and} \bibinfo{person}{et al}.}
  \bibinfo{year}{2019}\natexlab{}.
\newblock \showarticletitle{Libra {R-CNN:} Towards Balanced Learning for Object
  Detection}. In \bibinfo{booktitle}{\emph{{CVPR}}}. \bibinfo{pages}{821--830}.
\newblock


\bibitem[Qiao et~al\mbox{.}(2021)]%
        {DBLP:conf/iccv/QiaoZLQWZ21}
\bibfield{author}{\bibinfo{person}{Limeng Qiao}, \bibinfo{person}{Yuxuan Zhao},
  \bibinfo{person}{Zhiyuan Li}, \bibinfo{person}{Xi Qiu},
  \bibinfo{person}{Jianan Wu}, {and} \bibinfo{person}{Chi Zhang}.}
  \bibinfo{year}{2021}\natexlab{}.
\newblock \showarticletitle{DeFRCN: Decoupled Faster {R-CNN} for Few-Shot
  Object Detection}. In \bibinfo{booktitle}{\emph{{ICCV}}}.
  \bibinfo{pages}{8661--8670}.
\newblock


\bibitem[Rasmussen and Williams(2006)]%
        {DBLP:books/lib/RasmussenW06}
\bibfield{author}{\bibinfo{person}{Carl~Edward Rasmussen} {and}
  \bibinfo{person}{Christopher K.~I. Williams}.}
  \bibinfo{year}{2006}\natexlab{}.
\newblock \bibinfo{booktitle}{\emph{Gaussian Processes for Machine Learning}}.
\newblock \bibinfo{publisher}{{MIT} Press}.
\newblock


\bibitem[Ravi and Larochelle(2017)]%
        {DBLP:conf/iclr/RaviL17}
\bibfield{author}{\bibinfo{person}{Sachin Ravi} {and} \bibinfo{person}{Hugo
  Larochelle}.} \bibinfo{year}{2017}\natexlab{}.
\newblock \showarticletitle{Optimization as a Model for Few-Shot Learning}. In
  \bibinfo{booktitle}{\emph{{ICLR}}}.
\newblock


\bibitem[Razavian et~al\mbox{.}(2014)]%
        {DBLP:conf/cvpr/RazavianASC14}
\bibfield{author}{\bibinfo{person}{Ali~Sharif Razavian},
  \bibinfo{person}{Hossein Azizpour}, \bibinfo{person}{Josephine Sullivan},
  {and} \bibinfo{person}{Stefan Carlsson}.} \bibinfo{year}{2014}\natexlab{}.
\newblock \showarticletitle{{CNN} Features Off-the-Shelf: An Astounding
  Baseline for Recognition}. In \bibinfo{booktitle}{\emph{{CVPR} Workshops}}.
  \bibinfo{pages}{512--519}.
\newblock


\bibitem[Redmon et~al\mbox{.}(2015)]%
        {DBLP:journals/corr/RedmonDGF15}
\bibfield{author}{\bibinfo{person}{Joseph Redmon},
  \bibinfo{person}{Santosh~Kumar Divvala}, \bibinfo{person}{Ross~B. Girshick},
  {and} \bibinfo{person}{Ali Farhadi}.} \bibinfo{year}{2015}\natexlab{}.
\newblock \showarticletitle{You Only Look Once: Unified, Real-Time Object
  Detection}.
\newblock \bibinfo{journal}{\emph{CoRR}}  \bibinfo{volume}{abs/1506.02640}
  (\bibinfo{year}{2015}).
\newblock


\bibitem[Redmon and Farhadi(2018)]%
        {DBLP:journals/corr/abs-1804-02767}
\bibfield{author}{\bibinfo{person}{Joseph Redmon} {and} \bibinfo{person}{Ali
  Farhadi}.} \bibinfo{year}{2018}\natexlab{}.
\newblock \showarticletitle{YOLOv3: An Incremental Improvement}.
\newblock \bibinfo{journal}{\emph{CoRR}}  \bibinfo{volume}{abs/1804.02767}
  (\bibinfo{year}{2018}).
\newblock


\bibitem[Ren et~al\mbox{.}(2018)]%
        {DBLP:conf/iclr/RenTRSSTLZ18}
\bibfield{author}{\bibinfo{person}{Mengye Ren}, \bibinfo{person}{Eleni
  Triantafillou}, \bibinfo{person}{Sachin Ravi}, \bibinfo{person}{Jake Snell},
  \bibinfo{person}{Kevin Swersky}, \bibinfo{person}{Joshua~B. Tenenbaum},
  \bibinfo{person}{Hugo Larochelle}, {and} \bibinfo{person}{Richard~S. Zemel}.}
  \bibinfo{year}{2018}\natexlab{}.
\newblock \showarticletitle{Meta-Learning for Semi-Supervised Few-Shot
  Classification}. In \bibinfo{booktitle}{\emph{{ICLR}}}.
\newblock


\bibitem[Ren et~al\mbox{.}(2015)]%
        {DBLP:conf/nips/RenHGS15}
\bibfield{author}{\bibinfo{person}{Shaoqing Ren}, \bibinfo{person}{Kaiming He},
  \bibinfo{person}{Ross~B. Girshick}, {and} \bibinfo{person}{Jian Sun}.}
  \bibinfo{year}{2015}\natexlab{}.
\newblock \showarticletitle{Faster {R-CNN:} Towards Real-Time Object Detection
  with Region Proposal Networks}. In \bibinfo{booktitle}{\emph{{NIPS}}}.
  \bibinfo{pages}{91--99}.
\newblock


\bibitem[Salakhutdinov et~al\mbox{.}(2012)]%
        {DBLP:journals/jmlr/SalakhutdinovTT12}
\bibfield{author}{\bibinfo{person}{Ruslan Salakhutdinov},
  \bibinfo{person}{Joshua~B. Tenenbaum}, {and} \bibinfo{person}{Antonio
  Torralba}.} \bibinfo{year}{2012}\natexlab{}.
\newblock \showarticletitle{One-Shot Learning with a Hierarchical Nonparametric
  Bayesian Model}. In \bibinfo{booktitle}{\emph{{ICML}}}.
  \bibinfo{pages}{195--206}.
\newblock


\bibitem[Schwartz et~al\mbox{.}(2018)]%
        {DBLP:conf/nips/SchwartzKSHMKFG18}
\bibfield{author}{\bibinfo{person}{Eli Schwartz}, \bibinfo{person}{Leonid
  Karlinsky}, \bibinfo{person}{Joseph Shtok}, \bibinfo{person}{Sivan Harary},
  \bibinfo{person}{Mattias Marder}, \bibinfo{person}{Abhishek Kumar},
  \bibinfo{person}{Rog{\'{e}}rio~Schmidt Feris}, \bibinfo{person}{Raja Giryes},
  {and} \bibinfo{person}{Alexander~M. Bronstein}.}
  \bibinfo{year}{2018}\natexlab{}.
\newblock \showarticletitle{Delta-encoder: An Effective Sample Synthesis Method
  for Few-shot Object Recognition}. In \bibinfo{booktitle}{\emph{NIPS}}.
  \bibinfo{pages}{2850--2860}.
\newblock


\bibitem[Shen et~al\mbox{.}(2021)]%
        {DBLP:conf/cvpr/ShenLQ0CS21}
\bibfield{author}{\bibinfo{person}{Zhiqiang Shen}, \bibinfo{person}{Zechun
  Liu}, \bibinfo{person}{Jie Qin}, \bibinfo{person}{Lei Huang},
  \bibinfo{person}{Kwang{-}Ting Cheng}, {and} \bibinfo{person}{Marios
  Savvides}.} \bibinfo{year}{2021}\natexlab{}.
\newblock \showarticletitle{{S2-BNN:} Bridging the Gap Between Self-Supervised
  Real and 1-Bit Neural Networks via Guided Distribution Calibration}. In
  \bibinfo{booktitle}{\emph{{CVPR}}}. \bibinfo{pages}{2165--2174}.
\newblock


\bibitem[Snell et~al\mbox{.}(2017)]%
        {DBLP:conf/nips/SnellSZ17}
\bibfield{author}{\bibinfo{person}{Jake Snell}, \bibinfo{person}{Kevin
  Swersky}, {and} \bibinfo{person}{Richard~S. Zemel}.}
  \bibinfo{year}{2017}\natexlab{}.
\newblock \showarticletitle{Prototypical Networks for Few-shot Learning}. In
  \bibinfo{booktitle}{\emph{{NIPS}}}. \bibinfo{pages}{4077--4087}.
\newblock


\bibitem[Song et~al\mbox{.}(2019)]%
        {DBLP:conf/icml/SongDKF19}
\bibfield{author}{\bibinfo{person}{Hao Song}, \bibinfo{person}{Tom Diethe},
  \bibinfo{person}{Meelis Kull}, {and} \bibinfo{person}{Peter~A. Flach}.}
  \bibinfo{year}{2019}\natexlab{}.
\newblock \showarticletitle{Distribution Calibration for Regression}. In
  \bibinfo{booktitle}{\emph{{ICML}}}. \bibinfo{pages}{5897--5906}.
\newblock


\bibitem[Sun et~al\mbox{.}(2021)]%
        {DBLP:conf/cvpr/SunLCYZ21}
\bibfield{author}{\bibinfo{person}{Bo Sun}, \bibinfo{person}{Banghuai Li},
  \bibinfo{person}{Shengcai Cai}, \bibinfo{person}{Ye Yuan}, {and}
  \bibinfo{person}{Chi Zhang}.} \bibinfo{year}{2021}\natexlab{}.
\newblock \showarticletitle{{FSCE:} Few-Shot Object Detection via Contrastive
  Proposal Encoding}. In \bibinfo{booktitle}{\emph{{CVPR}}}.
  \bibinfo{pages}{7352--7362}.
\newblock


\bibitem[Sung et~al\mbox{.}(2018)]%
        {DBLP:conf/cvpr/SungYZXTH18}
\bibfield{author}{\bibinfo{person}{Flood Sung}, \bibinfo{person}{Yongxin Yang},
  \bibinfo{person}{Li Zhang}, \bibinfo{person}{Tao Xiang},
  \bibinfo{person}{Philip H.~S. Torr}, {and} \bibinfo{person}{Timothy~M.
  Hospedales}.} \bibinfo{year}{2018}\natexlab{}.
\newblock \showarticletitle{Learning to Compare: Relation Network for Few-Shot
  Learning}. In \bibinfo{booktitle}{\emph{{CVPR}}}.
  \bibinfo{pages}{1199--1208}.
\newblock


\bibitem[Tian et~al\mbox{.}(2019)]%
        {DBLP:conf/iccv/TianSCH19}
\bibfield{author}{\bibinfo{person}{Zhi Tian}, \bibinfo{person}{Chunhua Shen},
  \bibinfo{person}{Hao Chen}, {and} \bibinfo{person}{Tong He}.}
  \bibinfo{year}{2019}\natexlab{}.
\newblock \showarticletitle{{FCOS:} Fully Convolutional One-Stage Object
  Detection}. In \bibinfo{booktitle}{\emph{{ICCV}}}.
  \bibinfo{pages}{9626--9635}.
\newblock


\bibitem[Triantafillou et~al\mbox{.}(2017)]%
        {DBLP:conf/nips/TriantafillouZU17}
\bibfield{author}{\bibinfo{person}{Eleni Triantafillou},
  \bibinfo{person}{Richard~S. Zemel}, {and} \bibinfo{person}{Raquel Urtasun}.}
  \bibinfo{year}{2017}\natexlab{}.
\newblock \showarticletitle{Few-Shot Learning Through an Information Retrieval
  Lens}. In \bibinfo{booktitle}{\emph{{NIPS}}}. \bibinfo{pages}{2255--2265}.
\newblock


\bibitem[Vinyals et~al\mbox{.}(2016)]%
        {DBLP:conf/nips/VinyalsBLKW16}
\bibfield{author}{\bibinfo{person}{Oriol Vinyals}, \bibinfo{person}{Charles
  Blundell}, \bibinfo{person}{Tim Lillicrap}, \bibinfo{person}{Koray
  Kavukcuoglu}, {and} \bibinfo{person}{Daan Wierstra}.}
  \bibinfo{year}{2016}\natexlab{}.
\newblock \showarticletitle{Matching Networks for One Shot Learning}. In
  \bibinfo{booktitle}{\emph{{NIPS}}}. \bibinfo{pages}{3630--3638}.
\newblock


\bibitem[Wang et~al\mbox{.}(2021)]%
        {DBLP:conf/mm/WangWLL21}
\bibfield{author}{\bibinfo{person}{Jiahao Wang}, \bibinfo{person}{Yunhong
  Wang}, \bibinfo{person}{Sheng Liu}, {and} \bibinfo{person}{Annan Li}.}
  \bibinfo{year}{2021}\natexlab{}.
\newblock \showarticletitle{Few-shot Fine-Grained Action Recognition via
  Bidirectional Attention and Contrastive Meta-Learning}. In
  \bibinfo{booktitle}{\emph{{ACM} Multimedia}}. \bibinfo{publisher}{{ACM}},
  \bibinfo{pages}{582--591}.
\newblock


\bibitem[Wang et~al\mbox{.}(2020)]%
        {DBLP:conf/icml/WangH0DY20}
\bibfield{author}{\bibinfo{person}{Xin Wang}, \bibinfo{person}{Thomas~E.
  Huang}, \bibinfo{person}{Joseph Gonzalez}, \bibinfo{person}{Trevor Darrell},
  {and} \bibinfo{person}{Fisher Yu}.} \bibinfo{year}{2020}\natexlab{}.
\newblock \showarticletitle{Frustratingly Simple Few-Shot Object Detection}. In
  \bibinfo{booktitle}{\emph{{ICML}}}. \bibinfo{pages}{9919--9928}.
\newblock


\bibitem[Wang et~al\mbox{.}(2018)]%
        {DBLP:conf/cvpr/WangGHH18}
\bibfield{author}{\bibinfo{person}{Yu{-}Xiong Wang}, \bibinfo{person}{Ross~B.
  Girshick}, \bibinfo{person}{Martial Hebert}, {and} \bibinfo{person}{Bharath
  Hariharan}.} \bibinfo{year}{2018}\natexlab{}.
\newblock \showarticletitle{Low-Shot Learning From Imaginary Data}. In
  \bibinfo{booktitle}{\emph{{CVPR}}}. \bibinfo{pages}{7278--7286}.
\newblock


\bibitem[Wang et~al\mbox{.}(2019)]%
        {DBLP:conf/iccv/WangRH19}
\bibfield{author}{\bibinfo{person}{Yu{-}Xiong Wang}, \bibinfo{person}{Deva
  Ramanan}, {and} \bibinfo{person}{Martial Hebert}.}
  \bibinfo{year}{2019}\natexlab{}.
\newblock \showarticletitle{Meta-Learning to Detect Rare Objects}. In
  \bibinfo{booktitle}{\emph{{ICCV}}}. \bibinfo{pages}{9924--9933}.
\newblock


\bibitem[Wu et~al\mbox{.}(2021)]%
        {DBLP:journals/corr/abs-2103-01077}
\bibfield{author}{\bibinfo{person}{Aming Wu}, \bibinfo{person}{Yahong Han},
  \bibinfo{person}{Linchao Zhu}, \bibinfo{person}{Yi Yang}, {and}
  \bibinfo{person}{Cheng Deng}.} \bibinfo{year}{2021}\natexlab{}.
\newblock \showarticletitle{Universal-Prototype Augmentation for Few-Shot
  Object Detection}.
\newblock \bibinfo{journal}{\emph{CoRR}}  \bibinfo{volume}{abs/2103.01077}
  (\bibinfo{year}{2021}).
\newblock


\bibitem[Wu et~al\mbox{.}(2020b)]%
        {DBLP:conf/eccv/WuL0W20}
\bibfield{author}{\bibinfo{person}{Jiaxi Wu}, \bibinfo{person}{Songtao Liu},
  \bibinfo{person}{Di Huang}, {and} \bibinfo{person}{Yunhong Wang}.}
  \bibinfo{year}{2020}\natexlab{b}.
\newblock \showarticletitle{Multi-scale Positive Sample Refinement for Few-Shot
  Object Detection}. In \bibinfo{booktitle}{\emph{{ECCV}}}.
  \bibinfo{pages}{456--472}.
\newblock


\bibitem[Wu et~al\mbox{.}(2020a)]%
        {DBLP:conf/cvpr/0008CYLWL020}
\bibfield{author}{\bibinfo{person}{Yue Wu}, \bibinfo{person}{Yinpeng Chen},
  \bibinfo{person}{Lu Yuan}, \bibinfo{person}{Zicheng Liu},
  \bibinfo{person}{Lijuan Wang}, \bibinfo{person}{Hongzhi Li}, {and}
  \bibinfo{person}{Yun Fu}.} \bibinfo{year}{2020}\natexlab{a}.
\newblock \showarticletitle{Rethinking Classification and Localization for
  Object Detection}. In \bibinfo{booktitle}{\emph{{CVPR}}}.
  \bibinfo{pages}{10183--10192}.
\newblock


\bibitem[Xian et~al\mbox{.}(2018)]%
        {DBLP:conf/cvpr/XianLSA18}
\bibfield{author}{\bibinfo{person}{Yongqin Xian}, \bibinfo{person}{Tobias
  Lorenz}, \bibinfo{person}{Bernt Schiele}, {and} \bibinfo{person}{Zeynep
  Akata}.} \bibinfo{year}{2018}\natexlab{}.
\newblock \showarticletitle{Feature Generating Networks for Zero-Shot
  Learning}. In \bibinfo{booktitle}{\emph{{CVPR}}}.
  \bibinfo{pages}{5542--5551}.
\newblock


\bibitem[Xiao and Marlet(2020)]%
        {DBLP:conf/eccv/XiaoM20}
\bibfield{author}{\bibinfo{person}{Yang Xiao} {and} \bibinfo{person}{Renaud
  Marlet}.} \bibinfo{year}{2020}\natexlab{}.
\newblock \showarticletitle{Few-Shot Object Detection and Viewpoint Estimation
  for Objects in the Wild}. In \bibinfo{booktitle}{\emph{{ECCV}}}.
  \bibinfo{pages}{192--210}.
\newblock


\bibitem[Yan et~al\mbox{.}(2019)]%
        {DBLP:conf/iccv/YanCXWLL19}
\bibfield{author}{\bibinfo{person}{Xiaopeng Yan}, \bibinfo{person}{Ziliang
  Chen}, \bibinfo{person}{Anni Xu}, \bibinfo{person}{Xiaoxi Wang},
  \bibinfo{person}{Xiaodan Liang}, {and} \bibinfo{person}{Liang Lin}.}
  \bibinfo{year}{2019}\natexlab{}.
\newblock \showarticletitle{Meta {R-CNN:} Towards General Solver for
  Instance-Level Low-Shot Learning}. In \bibinfo{booktitle}{\emph{{ICCV}}}.
  \bibinfo{pages}{9576--9585}.
\newblock


\bibitem[Yang et~al\mbox{.}(2021b)]%
        {DBLP:conf/mm/0001YC21}
\bibfield{author}{\bibinfo{person}{Jinhai Yang}, \bibinfo{person}{Hua Yang},
  {and} \bibinfo{person}{Lin Chen}.} \bibinfo{year}{2021}\natexlab{b}.
\newblock \showarticletitle{Towards Cross-Granularity Few-Shot Learning:
  Coarse-to-Fine Pseudo-Labeling with Visual-Semantic Meta-Embedding}. In
  \bibinfo{booktitle}{\emph{{ACM} Multimedia}}. \bibinfo{publisher}{{ACM}},
  \bibinfo{pages}{3005--3014}.
\newblock


\bibitem[Yang et~al\mbox{.}(2021a)]%
        {DBLP:conf/iclr/YangLX21}
\bibfield{author}{\bibinfo{person}{Shuo Yang}, \bibinfo{person}{Lu Liu}, {and}
  \bibinfo{person}{Min Xu}.} \bibinfo{year}{2021}\natexlab{a}.
\newblock \showarticletitle{Free Lunch for Few-shot Learning: Distribution
  Calibration}. In \bibinfo{booktitle}{\emph{{ICLR}}}.
\newblock


\bibitem[Yang et~al\mbox{.}(2019)]%
        {DBLP:conf/iccv/YangLHWL19}
\bibfield{author}{\bibinfo{person}{Ze Yang}, \bibinfo{person}{Shaohui Liu},
  \bibinfo{person}{Han Hu}, \bibinfo{person}{Liwei Wang}, {and}
  \bibinfo{person}{Stephen Lin}.} \bibinfo{year}{2019}\natexlab{}.
\newblock \showarticletitle{RepPoints: Point Set Representation for Object
  Detection}. In \bibinfo{booktitle}{\emph{{ICCV}}}.
  \bibinfo{pages}{9656--9665}.
\newblock


\bibitem[Yu et~al\mbox{.}(2020)]%
        {DBLP:conf/cvpr/0004TLHWCJ020}
\bibfield{author}{\bibinfo{person}{Lu Yu}, \bibinfo{person}{Bartlomiej
  Twardowski}, \bibinfo{person}{Xialei Liu}, \bibinfo{person}{Luis Herranz},
  \bibinfo{person}{Kai Wang}, \bibinfo{person}{Yongmei Cheng},
  \bibinfo{person}{Shangling Jui}, {and} \bibinfo{person}{Joost van~de
  Weijer}.} \bibinfo{year}{2020}\natexlab{}.
\newblock \showarticletitle{Semantic Drift Compensation for Class-Incremental
  Learning}. In \bibinfo{booktitle}{\emph{{CVPR}}}.
  \bibinfo{pages}{6980--6989}.
\newblock


\bibitem[Zhang et~al\mbox{.}(2021a)]%
        {DBLP:conf/cvpr/ZhangLY0S21}
\bibfield{author}{\bibinfo{person}{Songyang Zhang}, \bibinfo{person}{Zeming
  Li}, \bibinfo{person}{Shipeng Yan}, \bibinfo{person}{Xuming He}, {and}
  \bibinfo{person}{Jian Sun}.} \bibinfo{year}{2021}\natexlab{a}.
\newblock \showarticletitle{Distribution Alignment: {A} Unified Framework for
  Long-Tail Visual Recognition}. In \bibinfo{booktitle}{\emph{{CVPR}}}.
  \bibinfo{pages}{2361--2370}.
\newblock


\bibitem[Zhang et~al\mbox{.}(2021b)]%
        {DBLP:conf/iclr/0007WKWA21}
\bibfield{author}{\bibinfo{person}{Tianyi Zhang}, \bibinfo{person}{Felix Wu},
  \bibinfo{person}{Arzoo Katiyar}, \bibinfo{person}{Kilian~Q. Weinberger},
  {and} \bibinfo{person}{Yoav Artzi}.} \bibinfo{year}{2021}\natexlab{b}.
\newblock \showarticletitle{Revisiting Few-sample {BERT} Fine-tuning}. In
  \bibinfo{booktitle}{\emph{{ICLR}}}.
\newblock


\bibitem[Zhang and Wang(2021)]%
        {DBLP:conf/cvpr/ZhangW21a}
\bibfield{author}{\bibinfo{person}{Weilin Zhang} {and}
  \bibinfo{person}{Yu{-}Xiong Wang}.} \bibinfo{year}{2021}\natexlab{}.
\newblock \showarticletitle{Hallucination Improves Few-Shot Object Detection}.
  In \bibinfo{booktitle}{\emph{{CVPR}}}. \bibinfo{pages}{13008--13017}.
\newblock


\bibitem[Zhao et~al\mbox{.}(2018)]%
        {DBLP:conf/eccv/ZhaoZYF18}
\bibfield{author}{\bibinfo{person}{Fang Zhao}, \bibinfo{person}{Jian Zhao},
  \bibinfo{person}{Shuicheng Yan}, {and} \bibinfo{person}{Jiashi Feng}.}
  \bibinfo{year}{2018}\natexlab{}.
\newblock \showarticletitle{Dynamic Conditional Networks for Few-Shot
  Learning}. In \bibinfo{booktitle}{\emph{{ECCV}}}. \bibinfo{pages}{20--36}.
\newblock


\bibitem[Zhou et~al\mbox{.}(2019)]%
        {DBLP:conf/cvpr/ZhouZK19}
\bibfield{author}{\bibinfo{person}{Xingyi Zhou}, \bibinfo{person}{Jiacheng
  Zhuo}, {and} \bibinfo{person}{Philipp Kr{\"{a}}henb{\"{u}}hl}.}
  \bibinfo{year}{2019}\natexlab{}.
\newblock \showarticletitle{Bottom-Up Object Detection by Grouping Extreme and
  Center Points}. In \bibinfo{booktitle}{\emph{{CVPR}}}.
  \bibinfo{pages}{850--859}.
\newblock


\bibitem[Zhu et~al\mbox{.}(2019)]%
        {DBLP:conf/cvpr/ZhuHS19}
\bibfield{author}{\bibinfo{person}{Chenchen Zhu}, \bibinfo{person}{Yihui He},
  {and} \bibinfo{person}{Marios Savvides}.} \bibinfo{year}{2019}\natexlab{}.
\newblock \showarticletitle{Feature Selective Anchor-Free Module for
  Single-Shot Object Detection}. In \bibinfo{booktitle}{\emph{{CVPR}}}.
  \bibinfo{pages}{840--849}.
\newblock


\end{thebibliography}

\appendix










\end{document}




\title{Supplementary Material for Exploring Effective Knowledge Transfer for Few-shot Object Detection}








\maketitle

\section{Ablation studies on {$\alpha$}}
In this section we give a simple experiment on parameter $\alpha$ in distribution calibration to demonstrate the effect of it. In distribution calibration, $\alpha$ is used in the calibration of biased novel class distribution:

\begin{equation}
    \sigma_n = \frac{1}{k} \sum_{c}^{|C_b^{cal}|} \sigma_c + \alpha
\label{e6}
\end{equation}

In this manner, $\alpha$ is a constant added to the calibrated covariance matrix and the function of $\alpha$ is controlling the degree of dispersion of generated features. The greater $\alpha$ is, the more scattered the features are. To explore the influence that $\alpha$ has on distribution calibration, we conduct a simple experiment. Experiment is conducted under 1-shot setting on split1 of PASCAL VOC dataset. To fully investigate the influence $\alpha$ has on distribution calibration, only distribution calibration is tested~(without shift compensation and distribution regularization). $k$ is fixed to 5. 

\begin{figure}[htb]
    \centering
    \includegraphics[width=\columnwidth]{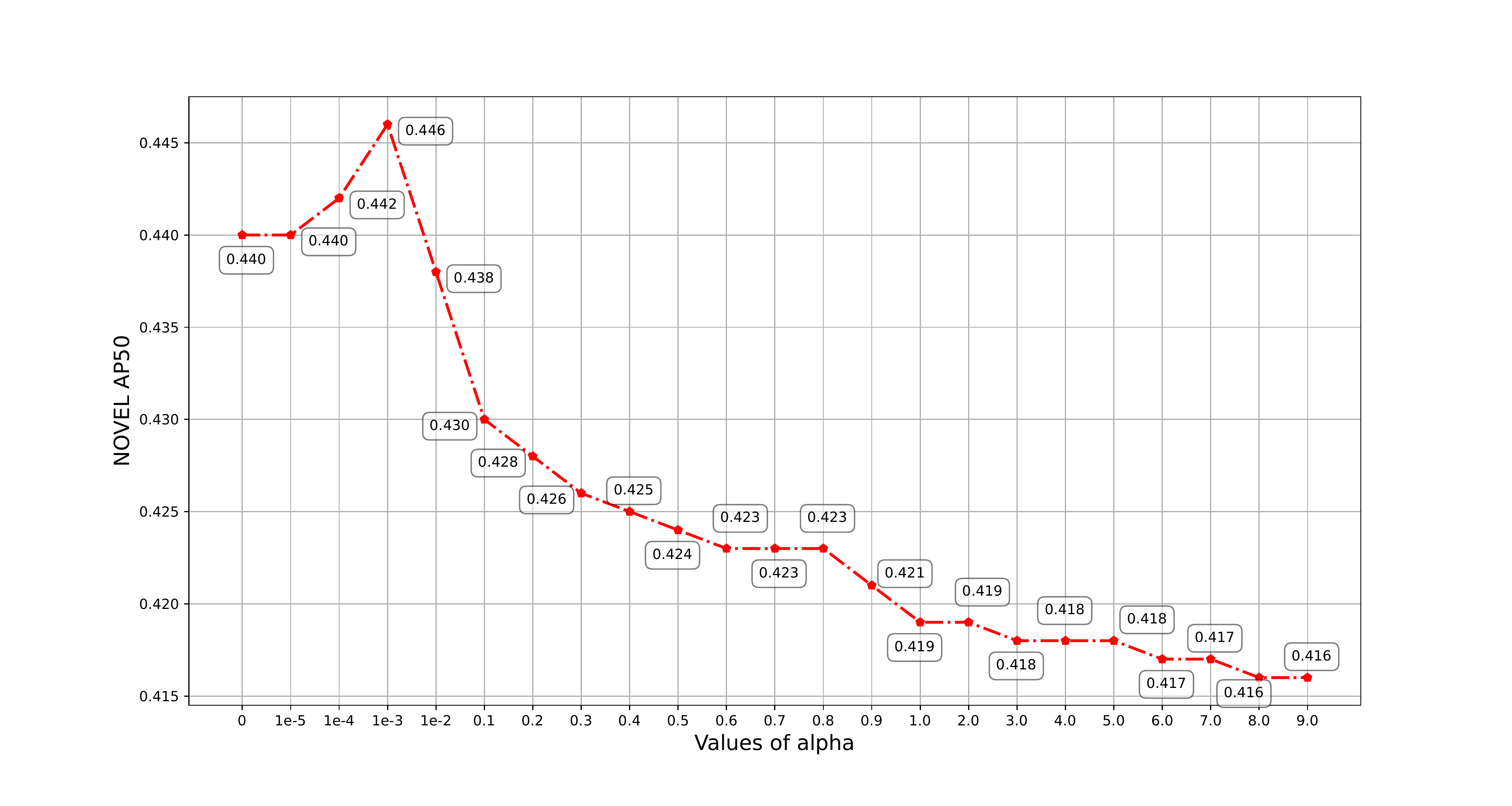}
    \caption{Ablation studies on $\alpha$.}
    \label{fig-alpha}
\end{figure}

Experimental results are shown in Figure~\ref{fig-alpha}. From the experimental results, we can see that (1) using $\alpha$ can lead to better performance than not using it. (2) when the value of $\alpha$ is too high, the performance of distribution calibration will be damaged, even inferior to not using $\alpha$. 

\section{Detection Results Visualization}
\begin{figure}[b]
    \centering
    \includegraphics[width=\columnwidth]{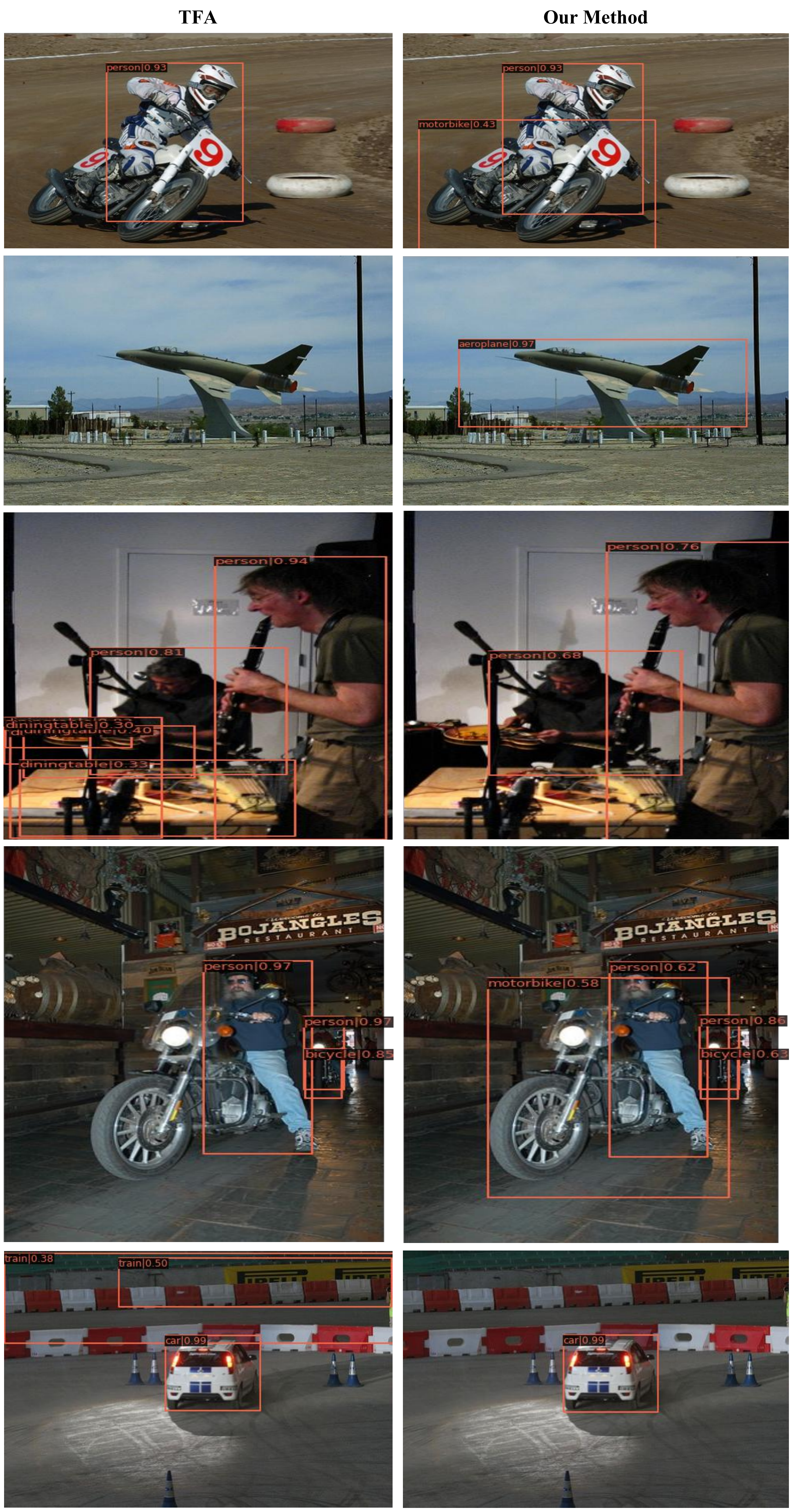}
    \caption{Visualization results of our proposed method and baseline TFA. Left: TFA right: our method.}
    \label{fig-demo}
\end{figure}

In this section, we provide qualitative visualizations of the detected novel objects on the split1 of PASCAL VOC, as shown in Figure~\ref{fig-demo}. Except from the results of our proposed method, we give visualization results of baseline TFA for comparison. Both our method and TFA are trained under 1-shot scenario. From visualization results, we can see that our method has stronger ability in detecting novel objects. Meanwhile, our method produces less false positives.

